\documentclass[letterpaper]{article} % DO NOT CHANGE THIS
\usepackage{aaai25}  % DO NOT CHANGE THIS
\usepackage{times}  % DO NOT CHANGE THIS
\usepackage{helvet}  % DO NOT CHANGE THIS
\usepackage{courier}  % DO NOT CHANGE THIS
\usepackage[hyphens]{url}  % DO NOT CHANGE THIS
\usepackage{graphicx} % DO NOT CHANGE THIS
\usepackage{natbib}  % DO NOT CHANGE THIS AND DO NOT ADD ANY OPTIONS TO IT
\usepackage{caption} % DO NOT CHANGE THIS AND DO NOT ADD ANY OPTIONS TO IT
\usepackage{algorithm}
\usepackage{algorithmic}

\usepackage{times}
\usepackage{soul}
\usepackage{url}
\usepackage{graphicx}
\usepackage{amsmath}
\usepackage{amsthm}
\usepackage{booktabs}
\usepackage{algorithm}
\usepackage{algorithmic}
\usepackage{multicol}
\usepackage{multirow}
\usepackage{graphicx}
\usepackage{amsthm}
\usepackage{csquotes}
\usepackage{tikz,tcolorbox}
\usepackage{listings}
\usepackage{enumitem}

\newtheorem{definition}{Definition}

\usepackage[utf8]{inputenc} % allow utf-8 input
\usepackage[T1]{fontenc}    % use 8-bit T1 fonts
\usepackage{url}            % simple URL typesetting
\usepackage{booktabs}       % professional-quality tables
\usepackage{amsfonts}       % blackboard math symbols
\usepackage{nicefrac}       % compact symbols for 1/2, etc.
\usepackage{microtype}      % microtypography
\usepackage{xcolor}         % colors

\newcommand{\model}{TrustUQA}
\usepackage{newfloat}
\usepackage{listings}
\DeclareCaptionStyle{ruled}{labelfont=normalfont,labelsep=colon,strut=off} % DO NOT CHANGE THIS
\floatstyle{ruled}
\newfloat{listing}{tb}{lst}{}
\floatname{listing}{Listing}
\title{TrustUQA: A Trustful Framework for Unified Structured Data Question Answering}
\author{
    Wen Zhang\textsuperscript{\rm 1,\rm 4}, Long Jin\textsuperscript{\rm 1}, Yushan Zhu\textsuperscript{\rm 1}, Jiaoyan Chen\textsuperscript{\rm 2}, Zhiwei Huang\textsuperscript{\rm 1}, Junjie Wang\textsuperscript{\rm 1}, Yin Hua\textsuperscript{\rm 1}\\, Lei Liang\textsuperscript{\rm 3}, Huajun Chen\textsuperscript{\rm 1,\rm 4,\rm 5}\thanks{Corresponding Author}
}
\affiliations{
    \textsuperscript{\rm 1}Zhejiang University \\
    \textsuperscript{\rm 2}University of Manchester, \textsuperscript{\rm 3}Ant Group \\
    \textsuperscript{\rm 4}ZJU-Ant Group Joint Lab of Knowledge Graph \\
    \textsuperscript{\rm 5}Zhejiang Key Laboratory of Big Data Intelligent Computing \\
    \{zhang.wen, longjin, yushanzhu, huajunsir\}@zju.edu.cn
    
}

\usepackage{bibentry}
\begin{document}

\maketitle

\begin{abstract}
Natural language question answering (QA) over structured data sources such as tables and knowledge graphs have been widely investigated, especially with Large Language Models (LLMs) in recent years. The main solutions include question to formal query parsing and retrieval-based answer generation. However, current methods of the former often suffer from weak generalization, failing to dealing with multi-types of sources, while the later is limited in trustfulness. In this paper, we propose {\model}, a trustful QA framework that can simultaneously support multiple types of structured data in a unified way. To this end, it adopts an LLM-friendly and unified knowledge representation method called \textit{Condition Graph} (CG), and uses an LLM and demonstration-based two-level method for CG querying. For enhancement, it is also equipped with dynamic demonstration  retrieval. We have evaluated {\model} with 5 benchmarks covering 3 types of structured data. It outperforms 2 existing unified structured data QA methods. In comparison with the baselines that are specific to one data type, it achieves state-of-the-art on 2 of the datasets. Further more,  we have demonstrated the potential of our method for more general QA tasks, QA over mixed structured data and QA across structured data. The code is available at https://github.com/zjukg/TrustUQA. 
% The code is available at https://anonymous.4open.science/r/TrustUQA . 
\end{abstract}

\section{Introduction}
% \vspace{-3mm}
% Structure Knoweldge representation and query(Symbolic) accurate but require background knowledge of query language 
%Knowledge representation and reasoning is one of the key topics towards achieving artificial intelligence. Many structured knowledge representation and querying methods have been proposed to represent and access the structured knowledge in the past decades, such as knowledge graph with SPARQL query language and table with SQL query language.
%However, people wanting to access the structured knowledge might not be familiar with the query language. 
%Accessing the structured knowledge via natural language (NL) is a more flexible solution, resulting a series of tasks called structured knowledge question answering (QA) including table QA \cite{tablegpt}, knowledge graph QA \cite{DBLP:conf/emnlp/SunDZMSC18}, temporal knowledge QA \cite{DBLP:conf/ijcai/LiuLLGLWW0FG23}, etc. 
%Many  NL2SQL  \cite{DBLP:journals/pvldb/KimSHL20} and NL2SPARQL  \cite{DBLP:journals/jiis/JungK20} works are proposed for structured knowledge QA by translating the natural language questions to structured queries (NL2SQ) and get the answer by executing queries.

Question answering (QA) seeking answers for a natural language question from structured data has attracted increasing attention in the past decade \cite{zhang2023survey,hu2023empirical}, leading to a few directions including QA over tables \cite{tablegpt}, QA over knowledge graph 
% which represents graph structured data composed of relation facts 
(i.e., KGQA) \cite{DBLP:conf/emnlp/SunDZMSC18}, QA over temporal KG \cite{DBLP:conf/ijcai/LiuLLGLWW0FG23} and so on.
A straightforward and widely studied solution is parsing questions to formal queries that can be executed on data storage and reasoning engines (a.k.a. NL2Query).
According to the data source types, these works are divided into NL2SQL  \cite{DBLP:journals/pvldb/KimSHL20} for relational databases, NL2SPARQL  \cite{DBLP:journals/jiis/JungK20} for KGs, etc.
However, current NL2Query methods are specifically developed for one query language corresponding to one data type and cannot be applied to others. This significantly limits their generality and usage in real-world scenarios, especially when it is unknown in which data resource the answer lies in, or the answer relies on data sources of multiple types.

% Many works are proposed for NL2SQL  \cite{DBLP:journals/pvldb/KimSHL20} and NL2SPARQL  \cite{DBLP:journals/jiis/JungK20}, etc, to translate the natural language question to structured queries (NL2SQ) and get the answer by executing queries.

% \begin{figure}
%     \centering
%     \includegraphics[width=.6\textwidth]{IJCAI2024/figures/intro.pdf}
%     \caption{Comparison between StructGPT and NeSyGPT, where SK and QF denote structured knowledge and query function.}
%     \vspace{-3mm}
%     \label{fig:intro}
% \end{figure}

%Due to the difference of the query language grammars, most NL2SQ methods are proposed for a specific query language and lack of universality. For example, NL2SQL methods could not be applied to translate questions to SPARQL and querying data in the knowledge graph. However, the ultimate goal of the structured knowledge QA is to  enable users accessing the structured knowledge through natural language and unaware of the query grammar and knowledge representation means. Thus uniform methods to accessing different kinds of structured knowledge through natural language is worth to investigate. 

With the development of LLMs and Retrieval-augmented Generation (RAG), another solution, which first retrieves relevant evidences of the question from data sources and then generates the answer, has become more and more popular~\cite{DBLP:conf/emnlp/JiangZDYZW23,cheng2024call}.
By utilizing different retrieval techniques for different structured data, this solution can lead to more general methods.
For example, StructGPT  \cite{DBLP:conf/emnlp/JiangZDYZW23} iteratively retrieves evidences and feeds them into an LLM  for answer generation, supporting KG, table and relational database.
%Benefit from the general natural language understanding and reasoning capability of the LLMs, 
%Retrieve Augmented Generation (RAG) \cite{DBLP:conf/emnlp/JiangZDYZW23,cheng2024call} methods are proposed for uniform structured knowledge QA with LLM. They retrieve useful evidence from structured data and let the LLM generate the answer by including the evidence in the prompt text input to the LLM. 
%For example, StructGPT  \cite{DBLP:conf/emnlp/JiangZDYZW23}
% , as shown in Figure \ref{fig:intro}, 
% explored using LLMs building general framework for diverse structured knowledge QA. Authors proposed 
%is an iterative reading-then-reasoning framework
%and includes a set of interfaces for each type of structured knowledge, through which the knowledge related to the questions could be obtained and included in the prompts of the LLMs for answer generation. 
% StructGPT shows the potential of structured knowledge question answering framework, while such a solution has some limitations. 
However, this solution suffers from some trustfulness issues: \textit{(i)} the generated answer may be inconsistent with the original structured data due to the hallucination of LLMs \cite{DBLP:journals/corr/abs-2309-01219}, the insistence of LLMs' parametric knowledge \cite{xie2023adaptive} and the irrelevant evidences that are retrieved; \textit{(ii)} a list of evidences are exposed to the LLM, potentially leaking private data if a third-party LLM is used; \textit{(iii)} the evidences can provide some plausible citations to the answer but it is hard to provide high quality explanations.
In contrast, NL2Query can mostly avoid the issue of \textit{(i)} and \textit{(ii)} as only some meta data and prototypes need to be exposed to the LLM to generate the query instead of the answer, and the issue of \textit{(iii)} since the query and its reasoning procedure are both accessible, providing logically rigorous explanation to the answer. 
Therefore, in this work, we adopt NL2Query, and propose \textbf{a trustful and unified framework named TrustUQA supporting different types of structured data simultaneously.} 
There are two desiderata for TrustUQA. (1) It is expected to represent diverse structured data.  Thus we propose \textbf{Condition Graph} (CG) and corresponding techniques for translating tables, KGs and temporal KGs into a CG. (2) It can support effective querying over the data representation. Thus we propose a method called \textbf{Lwo-layer Function-based CG Query}, which firstly uses LLMs to write basic queries (i.e., LLM queries) based on the question, and then uses pre-defined rules to transform these LLM queries to queries that can be executed on the CG (i.e., execution queries) for the eventual answer.
%eventually returns the answer from the CG.
Functions of the LLM query are designed with simple vocabularies that are more understandable by LLMs, and thus this method can get higher accuracy using no fine-tuning but few-shot prompting.
%includes not only \textit{execution query functions} but also \textit{LLM query functions}. The queries executed over the CG to get the answer are composed by execution query functions. The execution query functions are translated from LLM query functions. 
%We define the name and the parameter of the LLM query functions following the preference of LLMs based on our experience. LLMs could learn to write the LLM query functions through few-shot prompting. 
% The LLM query functions are easy for LLM to learn by defining the function and parameter names following the preference of LLMs based on the experience. 
% To better in-context learning effects of the LLMs,  
We also propose a \textbf{dynamic demonstration retrieving} method to further improve the prompt quality for higher accuracy.

To evaluate the effectiveness of {\model}, we experiment it on 5 benchmarks covering 3 kinds of structured data, including WikiSQL
\cite{DBLP:journals/corr/abs-1709-00103} 
and WTQ 
\cite{wtq} 
for table QA, WebQSP 
\cite{DBLP:conf/acl/YihRMCS16} 
and MetaQA 
\cite{DBLP:conf/aaai/ZhangDKSS18} 
for KG, and CronQuestion 
\cite{DBLP:conf/acl/SaxenaCT20} 
for temporal KG.
We compare its results with existing RAG-based unified QA methods \cite{DBLP:conf/emnlp/JiangZDYZW23,cheng2024call}.
Results show that {\model} outperforms these RAG-based methods on WikiSQL and  WebQSP. It also achieves state-of-the-art on WebQSP and CronQuestions compared to the models that support only one type of structured data. 

% Compared with the existing RAG-based uniform structured QA methods \cite{DBLP:conf/emnlp/JiangZDYZW23,cheng2024call}, {\model} performs better on . 
% We also compare {\model} to the ad hoc methods  designed for one kind of structured knowledge. Results show that {\model} achieves the state-of-the-art on 4 out of 5 datasets. 

%More importantly, benefit from the uniform condition graph representation, we show {\model} has great potential for general structured knowledge QA. 
Besides the comprehensive model analysis, we further demonstrate the generality of {\model} by questions whose answers potentially rely on different data sources --- tables, KG and temporal KG.
It includes two tasks: \textit{QA over mixed structured data} where each answer relies on one of the given data sources but which source it relies on is not specified, and \textit{QA across structured data} where answering the question relies on data from more than one sources.
These tasks are close to real-world scenarios, but have not been explored.
%Firstly, {\model} has capability of \textit{mixed structured knowledge QA}, that is answering questions about structured knowledge from tables, knowledge graphs, and temporal knowledge graphs with them mixed up. This achieves the ultimate goal of structured knowledge QA that enable users accessing structured knowledge unaware of the knowledge representation means. 
%Secondly, {\model} has capability of \textit{cross structured knowledge QA}, that is answering question requiring structured knowledge from both tables, knowledge graphs, and temporal knowledge graphs.
%These two tasks haven't been explored before due to the lack of methods with uniform representation. We believe these two QA tasks are worth to investigate in the future to achieve more general structured knowledge QA. 

In summary, our contributions lie in three aspects: (a) We propose a trustful framework {\model} for unified QA over multiple kinds of structured data. (b) We conduct comprehensives experiments over 5 benchmarks covering 3 types of structured data, proving the effectiveness of {\model}. (c) We demonstrate the generality of {\model} on two practical tasks that have never been explored before --- QA over mixed structured data and QA across structured data.

% \begin{itemize}[leftmargin=0.2cm, itemindent=0.2cm]
% \item We propose a trustful framework {\model} for unified QA over multiple kinds of structured data.
% It includes Condition Graph for general and LLM-friendly data representation, an LLM and demonstration-based two-level condition graph querying method, and a dynamic demonstration retrieval method for performance augmentation. 
% \item We conduct comprehensives experiments over 5 benchmarks covering 3 types of structured data, proving the effectiveness of {\model}. It outperforms the existing RAG-based QA methods and achieves the state-of-the-art on 3 out of 5 benchmarks.
% % We propose a unified and flexible disentangled structured knowledge querying framework, DISNEY, with LLMs, which applies LLM as the query writer and the CG data module as query executor.
% \item We demonstrate the generality of {\model} on two practical tasks that have never been explored before --- QA over mixed structured data and QA across structured data, both of which require the QA model to automatically get required data for a question from different given data sources.
% %which we think are worth to investigate in the future to achieve more general structured knowledge QA. 
% \end{itemize}

% \vspace{-3mm}
\section{Related Works}
% \vspace{-3mm}
% We classify existing methods of QA over structured data into two types: (1) NL2Query which translates questions to queries in formal languages and executes the queries for the answers, (2) generating the answer of the question without relying on the structured queries, i.e. NL2Answer. We first introduce these two types of methods and then introduce the unified methods. 
% Next, we introduce these two types of methods and discuss methods that use large language models.
% \subsection{NL2SQ-based Methods}
% \subsubsection{Structured Knowledge Representation}
% Representing knowledge in a structured way  \cite{levesque1986knowledge} to group related knowledge together and making it easier to query those knowledge has been explored for a long time. Many structured knowledge representation methods are proposed. 
% For example, knowledge with fixed schema is represented as tables, where the header includes the meaning of each column. 
% The fact knowledge expressing the relationship between two entities is represented as triples, and gathering triples together results in a labeled and directed graph, called knowledge graph  \cite{wikidata}. 
% More complex events with time stamps are represented as high-order tuples, such as quadruples and quintuples, and collecting them together forms a hypergraph  \cite{bretto2013hypergraph}. 

% % \subsection{Structured Knowledge Querying}
% \subsubsection{Structured Knowledge Query Generation}

\paragraph{NL2Query.} 
%%NL2Query methods translate questions to queries in formal languages and execute the queries for the answers.
% Given a question, a query of a formal language that exactly represents the meaning of the question is expected, and   
% the correct answer can be obtained by executing the query over the structured data. 
%%Since the query and the query execution procedure are accessible, they are relatively trustful. 
% These methods usually aim at QA over tables and KGs.
%
For table QA, the formal query language is SQL, and the task is also known as NL2SQL.
NL2SQL methods include schema-based approaches which build the query based on the schema of the database and the data indices  \cite{DBLP:conf/edbt/ZengLL16,DBLP:journals/tkde/LuoWLZWL11}, parsing-based approaches which parse the questions by understanding them through grammatical structures \cite{DBLP:conf/pldi/WangCB17,DBLP:journals/pacmpl/Yaghmazadeh0DD17}, and neural-machine-translation-based approaches which model NL2SQL as a language translation problem \cite{DBLP:conf/acl/GuoZGXLLZ19,DBLP:conf/acl/WangSLPR20}. 
% Pre-trained language models (PLMs), such as T5 \cite{t5} and BART \cite{bart}, with good language understanding ability are also used to NL2SQL tasks. 
% Pre-trained language models (PLMs) based methods
% % PLM-based methods 
% %DBLP:conf/emnlp/ScholakSB21 
% fine-tune PLMs on annotated training sets 
% % such as Spider  \cite{DBLP:conf/emnlp/YuZYYWLMLYRZR18} 
% for generating accurate SQL queries\cite{DBLP:conf/emnlp/QiTHW0ZWZL22,DBLP:conf/aaai/Li00023}.
Pre-trained language models (PLMs) based methods fine-tune PLMs to generate SQL queries~\cite{DBLP:conf/emnlp/QiTHW0ZWZL22,DBLP:conf/aaai/Li00023}, complete the SQL query sketch~\cite{DBLP:journals/corr/abs-2306-08891}, address the sub-queries \cite{DBLP:journals/corr/abs-2304-11015}. 
%with high effectiveness demonstrated on schema-alignment.
% For example, ZeroNL2SQL \cite{DBLP:journals/corr/abs-2306-08891} use the fine-tuned PLM to generate SQL sketches.  
% ZeroNL2SQL \cite{DBLP:journals/corr/abs-2306-08891} 
% % takes the advantages of fine-tuned PLMs 
% % (e.g. T5 \cite{t5}, BART \cite{bart}) 
% % on schema-alignment and LLMs on complex reasoning. Thus it 
% uses a fine-tuned PLM to generate a SQL query sketch and then applies LLMs to complete the query. DIN-SQL \cite{DBLP:journals/corr/abs-2304-11015} breaks down the SQL generation problem into sub-problems and uses LLMs to address each of them.
% such as detecting database tables and columns, 
%and feeds the solutions of those sub-problems into LLMs to generate the answer. 
We refer  \cite{DBLP:journals/vldb/KatsogiannisMeimarakisK23} for a conprehensive survey on NL2SQL. 

For KGQA, the formal query language is SPARQL, thus the task is also called NL2SPARQL. 
The methods usually include three steps: question understanding, linking and filtering  \cite{DBLP:journals/pacmmod/OmarDK023}. In  question understanding, some works \cite{DBLP:journals/tkde/Hu0YWZ18,DBLP:conf/acl/KapanipathiARRG21} 
% DBLP:conf/sigmod/ZouHWYHZ14
use natural language parsers such as dependency parsing while the others \cite{DBLP:journals/pacmmod/OmarDK023,DBLP:conf/acl/KapanipathiARRG21} use sequence to sequence generation to get graph patterns.
The linking step maps the mentions in questions to the entities and relations in the KG, where mapping dictionary  \cite{DBLP:conf/lrec/SpitkovskyC12}, indexing system  \cite{DBLP:conf/naacl/SakorMSSV0A19,DBLP:conf/semweb/DubeyBCL18}, embedding generation system \cite{DBLP:conf/emnlp/WuPJRZ20} are usually applied.
In filtering step, the type constraint is applied to filter answers.
LLMs are also applied for this task, in which generate-then-retrieve paradigm are usually applied that firstly generate the logical form and then bind the mentions to elements in KGs~\cite{DBLP:journals/corr/abs-2305-01750, DBLP:journals/corr/abs-2310-08975}. 
% for generating a draft of the questions, 
% KB-BINDER \cite{DBLP:journals/corr/abs-2305-01750}  uses an LLM to generate a draft of the question with few-shot prompting, and then binds the mentions to elements in the KG. 
% KB-BINDER-R retrieves the top-K similar examples in the training set as few-shot examples.

These NL2Query methods are developed towards one specific query language or even for one specific dataset, lacking generality for supporting different types of structured data.
%Thus they are not uniform frameworks for structured knowledge QA. 

% \subsection{Natural Language to Answer (NL2A)}

% \paragraph{Non-NL2SQ-based Structured Knowledge QA Methods.}
\paragraph{NL2Answer.}
% Different from NL2SQ2A, NL2A methods 
% Non-NL2SQ methods
NL2Answer methods
directly generate the answer, skipping the formal query.
%To achieve this, some works \cite{DBLP:conf/emnlp/SunDZMSC18} regard KGQA as multi-hop reasoning along with the relations until reaching the answer. 
For example, KGQA can be modeled as multi-hop reasoning \cite{DBLP:conf/emnlp/SunDZMSC18}.
Some works \cite{DBLP:conf/iclr/JiangZ0W23,DBLP:conf/emnlp/SunDZMSC18} trains a text encoder to encode the question and a reasoning module  to perform multi-hop reasoning. The others \cite{DBLP:conf/emnlp/JiangZZLW23} use a unified model for text encoding and reasoning. Since these models need to be trained, they are often only applicable to some specific datasets after training.
To address this problem, more general pre-training solutions that work across datasets are proposed. 
For example, TableGPT \cite{tablegpt} is a unified framework that enables LLMs to understand and operate on tables following natural languages. It pre-trains a general table encoder 
encoding tables into embeddings,
% to generate the table embedding, 
and prompt-tuning the LLM with table embedding as input over a vast corpus.
% comprising 2T tokens of textual data and 0.3M tables.
% The same as NL2Query methods, 
These methods are applicable to multiple datasets but cannot support different types of structured data.

% Thus in this paper, we are interested in more data-protectable and unified structured knowledge query methods.

\paragraph{Unified QA.} With LLMs, some unified QA frameworks that can support different types of structured data have been proposed.  
StructGPT  \cite{DBLP:conf/emnlp/JiangZDYZW23} 
% is a solution working for different kinds of structured knowledge. It 
% utilizes interfaces for different structured data to obtain relevant knowledge from the structured data and linearize and input them into LLM to directly generate the answer. 
is an iterative reading-then-reasoning framework. In reading, it collects evidence from the structured data through specialized functions for different kinds of structured data. In reasoning, it uses LLMs to generate the answer or the next reasoning step based on the collected evidence. 
Readi \cite{cheng2024call} is a reasoning-path-editing framework. Given a question, it first generates a reasoning path, and edits the path according to the feedback from instantiating the path over the structured data. It collects the KG evidence based on the edited reasoning paths and uses LLMs to generate the answer based on the evidence and the question. 
Though these two methods could be applied to different types of structured data, they are not unified enough, because the functions in StructGPT and the reasoning paths in Readi are specific to data types. 
% These two works are following the retrieve augmented generation (RAG) workflow that retrieve evidence from the structured data as input it to the LLM to help generate the answer. 
More importantly, they generate the answer based on evidence retrieval, and thus are not as trustful as NL2Query methods, as we discussed in Introduction.
%due to the necessary of inputting evidence data to the LLMs and the hallucination problem of LLMs, they are not as trustful as NL2Query methods. 

% 

% \vspace{-3mm}
\section{Methodology}
% \vspace{-3mm}
%%Our approach to achieve unified and trustful structured data QA relies on a general structured data representation whose queries can be accurately generated by LLMs according to the questions. 
%an uniform structured data representation and query languages, which allows to translate different structured data into the same representation form and use LLMs to generate accurate queries corresponding to the questions. 
{\model} uses Condition Graph (CG) for general and expressive data representation.
It can represent simple relationships, complex temporal facts, and rules.
For composing execution queries over CG, we design execution query function e.g., $search\_node()$ and $search\_condition()$.
%With CG, we designed execution query functions, for example $search\_node()$ and $search\_condition()$, as compositions of an execution query over CG. 
%Although writing execution queries with LLMs is challenging due to the novelty of CG, we simplify the process by having LLMs write more understandable functions, like $search\_information()$, using familiar concepts such as head entity, tail entity, relation, key, and value as inputs. 
As directly writing execution queries by LLMs is challenging due to their complexity and novelty, we propose a querying method called \textit{Two-layer Function-based CG Query}.
In the first layer, it uses LLMs to write simplified functions like $get\_information()$, whose vocabularies (head entity, tail entity, relation, key and value) are general and more understandable to LLMs, and applies such LLM functions for composing LLM queries.
In the second layer, it translates each LLM query into an execution query according to predefined rules.
%We refer to these functions as LLM functions and the resulting queries composed of them as LLM queries.
% We translate the LLM query into execution query before querying following rules. We called this as the two-layer GC query function. 
%We then translate each LLM query into an execution query according to predefined rules. This approach, involving both LLM query functions and execution query functions, is called two-layer CG query function.
% To enable LLM learn to write LLM queries, we use few-shot prompting with a few demonstrations of LLM queries included in the prompt. Considering the best demonstration for different questions are  different, we propose a dynamic demonstration retriever to improve the quality of LLM queries. 
To augment LLMs for writing LLM queries, we apply few-shot prompting with a few demonstrations of question and LLM query pairs, and propose a dynamic demonstration retriever, since the optimal demonstration varies with each question.
% Since the best demonstration varies depending on the question, we propose a dynamic demonstration retriever to improve the quality of LLM queries.
% Since the optimal demonstration varies with each question, we propose a dynamic demonstration retriever to enhance the quality of LLM queries.

\begin{figure}
    \centering
    % \vspace{-3mm}
\includegraphics[width=0.5\textwidth]{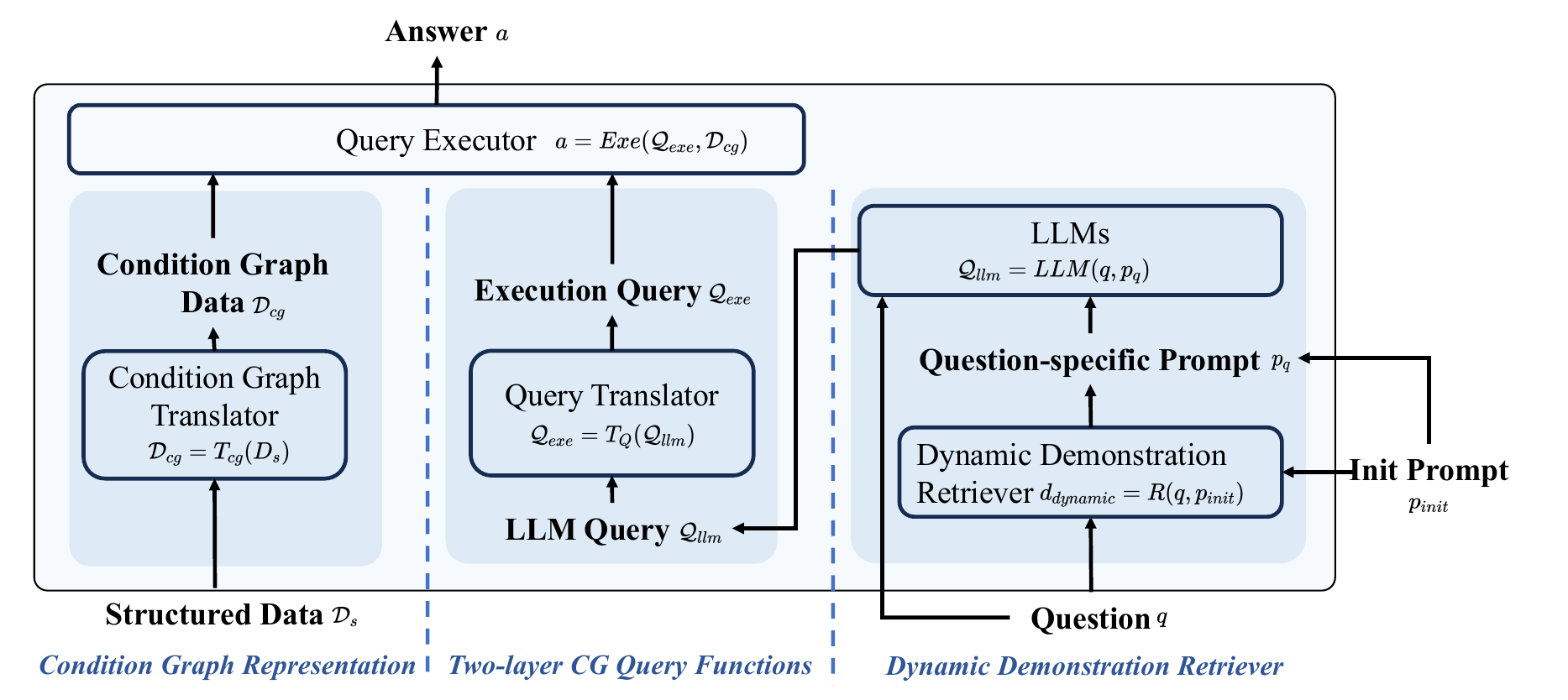}
\vspace{-4mm}
    \caption{Overview of the {\model} framework}
    \label{fig:{\model}}
    \vspace{-3mm}
\end{figure}

% Based on this thought process, 
Briefly, {\model} consists of 3 main modules, as shown in Figure \ref{fig:{\model}}: (1) Condition Graph Translator ($T_{cg}$) transforming input structured data into a CG, (2) Query Translator ($T_q$) translating LLM queries to execution queries over the CG, (3) Dynamic Demonstration Retriever ($R$) selecting the most similar examples from the training data as demonstrations for writing LLM queries.

\subsection{Condition Graph Translator $T_{cg}$}
Given structured data $\mathcal{D}_{s}$, $T_{cg}$ translates it into a condition graph which is formally defined below.
%a new structured data representation we proposed in this paper. 
% It is a uniform and expressive methods that could represent diverse structured data in the same way. 
% Following is the definition of the condition graph.

\begin{definition}
\textbf{Condition Graph} is a labeled directed graph represented as   
$\mathcal{CG}=\{\mathcal{N}, \mathcal{T}\}$. 
$\mathcal{N}$ is the set of nodes in CG.  Each node is combined with a string to represent the semantic meaning, which could be entities such as \texttt{Earth}, relationships such as \texttt{has friends}, properties such as \texttt{time}, or numerical values such as \texttt{2024}. 
% $\mathcal{N} \in \mathcal{E}$ is the set of nodes in the condition graph.
$\mathcal{T}=\{ (node_1, node_2, condition)| node_1 \in \mathcal{N}, node_2 \in \mathcal{N}\}$ is a collection of condition triples representing an edge that $node_1$ is connected to  $node_2$ under the $condition$, where $condition = [node_a, node_b, ...] $ is a list of nodes in $\mathcal{N}$. 
% The $condition$ is allowed to be empty ([]) which means $node_1$ is connected to  $node_2$ without condition.
The $condition$ can be empty ([]) indicating that $node_1$ is connected to $node_2$ without any conditions.
\end{definition}

Condition graph is expressive. The triple in the CG can represent (1) simple relationships such as (\texttt{Born In}, \texttt{Ulm}, [\texttt{Albert Einstein}]), (2) complex facts such as (\texttt{date}, \texttt{14 March 1879}, [\texttt{Albert Einstein}, \texttt{Born In}, \texttt{Ulm}]), (3) common rules such as (\texttt{Person}, \texttt{Has Parents}, []), meaning that each person has parents. 

Condition graph is a unified representation. 
% For different types of structured data, the condition graph translator $T_{cg}$ translates them following different rules. 
Next, we introduce how to translate three typical structured data --- tables, KGs, and temporal KGs, into a CG.

\paragraph{Translating Tables.}
% Each table in $\mathcal{D}_s$ is represented as 
A table as $\mathcal{D}_s$ is represented as
% \paragraph{For $\mathcal{D}_{sk} = Table$.}
$Table = \{ R, C, V\}$ which contains $m$ rows $R = \{row_i\}_{i=1}^{m}$ and $n$ columns $C =\{col_j\}_{j=1}^{n}$. The table header $row_1$ records the semantic meaning of each column. Starting from the second row, each row is a record. $V = \{v_{i,j} | i\in [1, m], j \in [1, n] \}$ represents the cell values of the table.
For example, a table with \textit{ name | born in city | Time  } as the header and \textit{ Albert Einstein | Ulm | 14 March 1879} as the record. 
During transformation, the translator $T_{cg}$ first inserts a column $c_0$ in front of the first column that $C := \left\{c_0, col_1, ..., col_n\right\}$.
% $C =\{col_j\}_{j=1}^{n+1}$. 
The values of the inserted column cells are strings indicating the order of the corresponding row, i.e. for the $i$th row, the inserted value is $[line\_i]$. 
% Starting from the second row,  $T_{cg}$ generates two condition triples for each value in the row.
For example, after insertion, the previous example table becomes \textit{ [line\_1]| name | born in city | Time  } as the header and \textit{ [line\_2] |  Albert Einstein | Ulm | 14 March 1879} as the record.
After inserting, 
starting from the second row and the second column, 
the translator will generate two condition triples for each cell. For example, for $v_{2,2}=\textit{Albert Einstein}$ in the example table, there will be (\texttt{ $[line\_2]$}, \texttt{name}, []) represents the entity $[line\_2]$ has name, and (\texttt{name}, \texttt{Albert Einstein}, $[line\_2]$) represents the name of entity $[line\_2]$ is Albert Einstein.
% $(\texttt{name}, \texttt{Albert Einstei}, [2])$, $\texttt{born in city}, \texttt{Ulm}, [2])$, $(\texttt{time}, \texttt{14 March 1879}, [2])$, $(\texttt{2}, \texttt{name}, [])$.
Formally, after translation,  
$\mathcal{D}_{cg}^{table} = \{ \mathcal{N}_{table}, \mathcal{T}_{table} \}$, where 
$\mathcal{N}_{table} = V \cup \{line\_1, line\_2, ..., line\_m\}$, 
$\mathcal{T}_{table} = \{([line\_i], v_{1,j}, []), (v_{1,j}, v_{i,j}, [line\_i]) | i \in [2, m], j \in [2, n+1]\}$.
 
% $\mathcal{N} = \{ v_{i,j} \cup \{i\}| i \in [1, m], j \in [1, n]\}$,
% $\mathcal{CG}_{table} = \{ \mathcal{N}_{table}, \mathcal{T}_{table}\}$ that $\mathcal{N}_{table} = \{ v_{i,j} \cup \{i\}| i \in [1, m], j \in [1, n]\}$, $\mathcal{T}_{table} = \{(i, v_{1,j}, []), (v_{1,j}, v_{i,j}, [i]) | i \in [2, m], j \in [1, n]\}$. 

\paragraph{Translating Knowledge Graph.}
% \paragraph{Translating knowledge graph into CG.}
A KG as $\mathcal{D}_{s}$ is represented as $\mathcal{KG}= \{ \mathcal{E}, \mathcal{R}, \mathcal{F}\}$, where $\mathcal{E}$, $\mathcal{R}$ and $\mathcal{F}$ are the set of entities, relations and  facts. $\mathcal{F}=\{ (h, r, t) | \{h,t\}\in\mathcal{E}, r\in \mathcal{R}\}$ is a set of triples representing the relations between entities. 
Translator $T_{cg}$ transforms a $\mathcal{KG}$ by generating two condition triples for each knowledge graph fact. 
For example, given a triple \textit{(Albert Einstein, born in city, Ulm)}, one condition triple (\texttt{Albert Einstein}, \texttt{born in}, []) denotes Albert Einstein has the property \textit{born in}, another condition triple (\texttt{born In}, \texttt{Ulm}, [\texttt{Albert Einstein}]) means if it is for Albert Einstein, the value of \textit{born in} is Ulm.
Thus after translation, 
$\mathcal{D}_{cg}^{kg} = \{\mathcal{N}_{kg}, \mathcal{T}_{kg} \}$, where $\mathcal{N}_{kg} = \mathcal{E}\cup\mathcal{R}$, and $\mathcal{T}_{kg} = \{(h, r, [ ]), (r, t, [h])| (h,r,t) \in \mathcal{F}\}$.
% that $\mathcal{CG}_{KG} =$ $ \{ \mathcal{N}_{KG}, \mathcal{T}_{KG}\}$, 
% where $\mathcal{N}_{KG} = \mathcal{E}\cup\mathcal{R}$, $\mathcal{T}_{KG} = \{(h, r, [ ]), (r, t, [h])| (h,r,t) \in \mathcal{T}\}$. 

\paragraph{Translating Temporal Knowledge Graph.}
% \paragraph{Translating temporal KGs into CG.}
A temporal KG as the $\mathcal{D}_s$ is represented as $\mathcal{TKG}=\{ \mathcal{E}, \mathcal{R}, \tau, \mathcal{Q}\}$, where $\mathcal{E}, \mathcal{R} $ and $\tau$ are the entity, relation, and time sets, and $\mathcal{Q} = \{ (h,r,t, \tau_s, \tau_e) \}$ is the quintuple set where $h, t \in \mathcal{E}$, $r \in \mathcal{R}$, and $\tau_s, \tau_e \in \tau$ denote the start time and end time of the fact $(h,r,t)$, respectively. 
% Given a temporal KG, 
% The translator $T_{cg}$ transforms a $\mathcal{TG}$ by generating at least 5 condition triples for each TKG triple. 
After translation, 
$\mathcal{T}_{tkg}$ collects all condition triples translated from each TKG triple. Specifically, 
for each $(h,r,t, \tau_s, \tau_e) $ in $\mathcal{Q}$, following condition triples are generated $(h,r,[ ])$, $(r,t,[h])$, $(\texttt{start time}, \tau_s, [h,r,t])$, $(\texttt{end time}, \tau_e, [h,r,t])$, $\{(\texttt{time}, \tau', [h,r,t] )| \tau' \in \{\tau_s, \tau_s+1, , \tau_s+2, ..., \tau_e\}\}$, where
$\texttt{start time}, \texttt{end time}$ and $\texttt{time}$ are three built-in terms introduced during translation and $\tau'$ are the integer time stamps between the start time and the end time. 
Thus after translation, 
$\mathcal{D}_{cg}^{tkg} = \{ \mathcal{N}_{tkg}, \mathcal{T}_{tkg} \}$. 
$\mathcal{N}_{tkg} = \mathcal{E}\cup \mathcal{R} \cup \tau \cup \{ \texttt{start time}, \texttt{end time}, \texttt{time}\} \cup \{\tau'\}$.
% where $\texttt{start time}, \texttt{end time}$ and $\texttt{time}$ are three built-in elements introduced during translation, and $\tau'$ are the integer time stamps between the start time and the end time. 

\begin{table*}[]
    \centering
    % \scriptsize
    \vspace{-3mm}
    \caption{Rules for translating LLM query functions to execution query functions. $gi$, $h$, $r$, $t$, $k$, $v$ denote $get\_information$, $head\_entity$, $relation$, $tail\_entity$, $key$, and $value$, $sn$, $sc$, $n1$, $n2$, $c$, and $s$ denote $search\_node$, $searh\_condition$, $node_1$, $node_2$, $condition$ and $scope$. For the parameters $t$ and $v$ of $get\_information$, $op$ is one of $\{\textgreater$, $\textless$, $=$, $\geq$, $\leq\}$ and is ``$=$'' in the table as example.}
    \vspace{-2mm}
    \resizebox{0.9\textwidth}{!}{
    \begin{tabular}{l|l }
    \toprule
      \textbf{LLM Query Function} & \textbf{Execution Query Function // Interpretation}  \\
     \midrule
     $gi(h=H)$ & $sn(n1=H)$ // return the relations and keys that the head entity $H$ has \\
     \hline 
     $gi(r=R)$ & $sn(n1=R)$ // return the tail entities under relation $R$\\
     \hline 
     $gi(k=K)$ & $sn(n1=K)$ // return the values under key $K$\\
     \hline 
     $gi(h=H, r=R)$ & $sn(n1=R, c=H)$ // return the tail entities of head entity $H$ under relation $R$\\
     \hline 
     $gi(h=H, k=K)$ & $sn(n1=K, c=H)$ // return the values of head entity $H$ under key $K$\\
     \hline 
     %$gi(h=H, t$``$=$''$T)$ & & return \\
     $gi(r=R, t$``$=$''$T)$ & $sc(n1=R, n2=T, op=$``$=$''$)$ // return the head entities whose tail entity is $T$ under relation $R$\\
     \hline 
     $gi(k=K, v$``$=$''$V)$ & $sc(n1=K, n2=V, op=$``$=$''$)$ // return the head entities whose value is $V$ under key $K$\\
     \hline
     \multirow{3}{*}{$gi(r=R, t$``$=$''$T, k=K, v$``$=$''$V)$}
      &$output\_of\_query_1 = sc(n1=R, n2=T, op=$``$=$''$)$ // return the head entities whose tail entity  \\ 
      & $output\_of\_query_2=sc(n1=K, n2=V, op=$``$=$''$)$ // is $T$ under relation $R$ and value is $V$  \\ 
      & $set\_interaction(set1=output\_of\_query_1, set2=output\_of\_query_2)$  // under key $K$\\
      \hline 
     \multirow{2}{*}{$gi(r=R, t$``$=$''$T, k=K)$} & $output\_of\_query_1 = sc(n1=R, n2=T, op=$``$=$''$)$ // return the value of the key $K$  
     \\
     & $sn(n1=K, s=output\_of\_query_1)$ // of the head entity whose tail entity is $T$ under relation $R$\\ 
     \hline 
     \multirow{2}{*}{$gi(r=R, k=K, v$``$=$''$V)$} & $output\_of\_query_1 = sc(n1=K, n2=V, op=$``$=$''$)$ // return the tail entity of the relation $R$  
     \\
     & $sn(n1=R, s=output\_of\_query_1)$ // of the head entity whose value is $V$ under key $K$\\ 
     \bottomrule
    \end{tabular}
    }
    % \vspace{-3mm}
    \label{tab:translation}
\end{table*}

\subsection{Query Translator $T_{Q}$}
% \vspace{-2mm}
The query translator $T_{Q}$ translates the LLM query $\mathcal{Q}_{llm}$ into the execution query $\mathcal{Q}_{exe}$ that can be executed over the CG.
% We designed two-level condition graph queries, the execution queries and LLM queries. This enables LLM learn to master the condition graph through prompting. 
% Next, we introduce execution query $\mathcal{Q}_{exe}$, the LLM query $\mathcal{Q}_{llm}$, and how to translate $\mathcal{Q}_{llm}$ to $\mathcal{Q}_{exe}$.
% Next, we introduce the LLM query $\mathcal{Q}_{llm}$, the execution query $\mathcal{Q}_{exe}$, and how to translate $\mathcal{Q}_{llm}$ to $\mathcal{Q}_{exe}$.

\paragraph{LLM Query $\mathcal{Q}_{llm}$.} The LLM query is  generated by an LLM corresponding to the question. 
% We realize that it is hard to teach an LLM to learn the new type of structured knowledge representation method and to write the execution query function through prompting methods. Enabling the LLM to write the execution query requires fine-tuning. 
% Thus we simplify and aggregate the execution query functions, resulting LLM query functions to compose the LLM query. 
% Specifically, we simplify the search functions into one LLM search function:
LLMs know the common vocabularies of head entity, relation, tail entity from KGs, as well as key, value used in tables and property graphs well.
% because they have been trained on such data. 
Thus we design the following \textit{searching function} to search information from the graph:
% \begin{align}
% \label{equ:get_information}
%     get\_information( head\_entity, relation,  \\
%     tail\_entity, key, value)
% \end{align}
\begin{equation}
\begin{aligned}
\label{equ:get_information}
    get\_information( & \, head\_entity, relation,  \\
    & \, tail\_entity, key, value)
\end{aligned}
\end{equation}
where the variables $head\_entity$, $relation$, $tail\_entity$, $key$ and $value$ are set to $None$ by default. 
% Different from condition graph, we find the LLM are familiar with the concept of head entity, relation, tail entity, key and value. Thus we make them as variables in $get\_information$ function. 
This function is able to represent complex queries. For example, $get\_information(head\_entity\text{=}None, relation\text{=}\texttt{Won},  $ $tail\_entity \text{=} \texttt{Nobel Prize}, key\text{=}\texttt{Year},$ $ value >\text{2000})$ represents "Who are the Nobel Prize Winners after 2000?"
% means searching entities that won Nobel Prize after 2000. 
 $get\_information( head\_entity\text{=}\texttt{Albert Einstein}, $ $relation\text{=}\texttt{Won},  tail\_entity \text{=} \texttt{Nobel Prize}, key\text{=}\texttt{Year}, $ $ value \text{=} None)$ 
 corresponds to "In which year did Albert Einstein won the Nobel Prize?"
 % means searching the year that Albert Einstein won the Nobel Prize. 

Apart from $get\_information()$, we also design a set of  \textit{reasoning functions} as follows:
% It is easy to understand through the function name.
\begin{itemize}[leftmargin=0.2cm, itemindent=0.2cm]
    \item {Set operations:} $set\_intersection(set_1, set_2)$, $set\_union$$(set_1$$,set_2)$,  $set\_difference(set_1, $ $set_2)$, $set\_negation(set_1, set_2)$, and $keep(set, value)$\footnote{An example is $keep(set=\{1,2,3\}, value<2)$}
    \item {Simple calculations:} $mean()$, $max()$, $min()$, $count()$.
\end{itemize}
%Based on these functions, complex LLM queries with multiple and nested searching and reasoning steps could be written. 
With these functions, multiple (nested) searching and reasoning steps can be expressed by LLM queries. We give examples of these reasoning functions in Appendix \ref{sec:examples4reasoning}.

\paragraph{Execution Query $\mathcal{Q}_{exe}$.} 
\label{sec:execution-query}
The execution query is executed over the CG, and is composed of a set of \textit{execution query functions} which include \textit{searching functions} and \textit{reasoning functions}. The reasoning functions are the same as the LLM reasoning functions. 
% Thus we mainly introduce the searching functions in this section.  
%
The search functions are used to obtain information from the CG following constrains:  
\begin{itemize}[leftmargin=0.2cm, itemindent=0.2cm]
    \item $search\_node(node_1, condition, scope)$: return $node_2$ of the condition triple $(node_1, ?, [condition])$ from the condition triple set denoted as $scope$. 
    % in $scope$ where $scope$ is a condition triple set.
     $condition\text{=}None$ means condition list is empty.
    $scope\text{=}all$ by default means including all the condition triples in $\mathcal{D}_{cg}$. For example 
    $search\_node(node1\text{=}\texttt{Born In},\; condition \text{=} \texttt{Albert  } $ $\texttt{Einstein}, scope\text{=}all)$ means getting the born-in information of Albert Einstein based on all condition triples. 
    \item $search\_condition(node_1, node_2\_value, op)$: 
    return $condition$ in the condition triple $(node_1, node_2, ?)$ such that $node_2$ satisfies the operation $op$ with respect to $node_2\_value$. For example, if $op$ is `$>$', the function return $condition$ where $node_2 > node_2\_value$. 
     Specifically, $search\_condition(node_1 \text{=} \texttt{Born}, \; node_2\_value\text{=} \texttt{2020}, $ $op\text{=}'>')$ means searching entities borning after 2020. 
     % The comparison between $node_2$ and $node_2\_value$  $compare(op, node_2\_value, node_2)$ returns $true$ 
     To determine if the $op$ constrain meets,  $compare(op, val_1, val_2)$ will be called.
     
    \item $compare(op, val_1, val_2)$: return $ture$ if $val_1$ $\{op\}$ $val_2$, otherwise return $false$, where $op$ is one of the comparison symbols $\{\textgreater$, $\textless$, $=$, $\geq$, $\leq\}$. 
    % The LLM function will be called if the comparison is not executed successfully.
    For example, $compare (op\text{=}'>', val_1 \text{=} 2024, val_2\text{=}2020)$  will return $true$.
    %because $2024>2020$. 
\end{itemize}
% The reasoning functions include the following. It is easy to understand through the function name.
% \begin{itemize}[leftmargin=0.2cm, itemindent=0.2cm]
%     \item {Set operations:} $set\_union$($set_1, set_2)$, $set\_intersection(set_1, set_2), set\_difference(set_1, $ $set_2)$, $set\_negation(set_1, set_2)$, and $keep(set, value)$\footnote{An example is $keep(set=\{1,2,3\}, value<2)$}
%     \item {Simple calculations:} $mean()$, $max()$, $min()$, $count()$.
% \end{itemize}

%Based on these functions, complex execution queries can be expressed. 
Given a CG $\mathcal{D}_{cg}$ and complex execution queries based on these function $\mathcal{Q}_{exe}$, the query executor $Exe$ can automatically execute $\mathcal{Q}_{exe}$ over $\mathcal{D}_{cg}$ to get the answer. 
%If the execution query follows the definition in Sec \ref{sec:execution-query}, it can be automatically executed. 

% By executing the execution query $\mathcal{Q}_{exe}$ over the condition graph data $\mathcal{D}_{cg}$, the query executor will output the answer. 

\paragraph{Translating $\mathcal{Q}_{llm}$ to $\mathcal{Q}_{exe}$.}
There are semantic mapping and syntax mapping steps for $T_{Q}$ to translate $\mathcal{Q}_{llm}$ to $\mathcal{Q}_{exe}$. (1) \textit{Semantic mapping}: mapping the  variable values in  $\mathcal{Q}_{llm}$ to the nodes in the $\mathcal{D}_{cg}$. To achieve this, we use a dense text encoder $E$, such as SentenceBert \cite{DBLP:conf/emnlp/ReimersG19}, to encode the nodes and variable values into vectors and map each value to the most similar node. We replace the value by the string of the mapped node in the function, resulting $\mathcal{Q}_{llm}'$. 
(2) \textit{Syntax mapping:} translating $\mathcal{Q}_{llm}'$ to $\mathcal{Q}_{exe}$ following a fixed set of rules. 
We summarize the translation rules in Table \ref{tab:translation}. 
There is a chance that the generated LLM queries are composed of undefined LLM query functions. For example, LLM might generate $compare(\textit{A's\ score}, \textit{B's\ score})$ for question ``\textit{For A and B, who has a larger score?}''. For such LLM query, the translator $T_{Q}$ keeps the query as it is during translation. And during execution, the \textit{LLM function} is called to use LLM to infer the output. Specifically, we include the function name and parameters in the prompt input to the LLM to generate the output. The LLM function prompt is shown in Appendix~\ref{sec:LLM_prompt}.

% \subsection{Query Executor $Exe$} 
 
\vspace{-2mm}
\subsection{Dynamic Demonstration Retriever $R$}
% \vspace{-2mm}
Given a question $q$, we use an LLM to generate the LLM query, denoted as $\mathcal{Q}_{llm} = LLM(q, p_q)$ where $p_q$ is another input acting as the question-specific prompt. $p_q$ includes a few question-query examples, called demonstrations. 
% Thus $p_q = p_{n} + p_{demo}$ where $p_{n}$ is the normal prompt text and  $p_{demo}$ is the few-shot prompt text. 
The initial demonstrations are manually crafted with a focus on representativeness and diversity, shown in Appendix~\ref{sec:prompt}.  
% While considering different questions, 

However, the optimal demonstration for different questions varies. For example, for question \textit{``What is the most common language in Norway?''}, the LLM query of the question \textit{``What is the major language spoken in Canada?''} is more informative than \textit{``Where the queen of Denmark lives?''}. 
% Thus we propose that dynamic demonstrations should be applied to achieve better performance. 
Thus we propose a dynamic demonstration retriever to retrieve $k$ most similar questions of $q$ from the training dataset $D_{train} 
\text{=} \{ (q_{train}, a_{train})\}$ as the demonstrations. 
% The dynamic retriever will rewrite the initial prompt $p$ into $p'$, denoted as $p'= R(q, p, q_{train})$.
Specifically, given a question $q$, we use a text encoder $E$ to encode $q$ and training question $q_{train}$ into vectors. Then we calculate the similarity of the question vectors and select the $m$ most similar training questions (denoted as $\mathcal{S}$) where $m>k$. 
% that is 
% $q' = arg\; top _{q_i \in q_{train}} m sim(E(q), E(q_i))$
% $\mathcal{S} = \underset{q_i \in q_{train}}{\text{arg top } m} \; sim(E(q), E(q_i)) $.
% ${q}' = Max_{m}(sim(E(q), E(q_{i}), q_{i}\in {q_{train}})$. 
% $q' = \operatorname{argmax}_{|I| = m \land I  \subseteq  q_{train}} \sum_{q_c \in I} sim(E(q), E(q_c)) $
Then we iteratively generate the LLM query $ \mathcal{Q}_{llm}^{q_i} \text{=} LLM(q_{i}, p_{q_i})$ from the most to the least similar $q_i \in \mathcal{S}$. If the result from $ Exe(T_{Q}(\mathcal{Q}_{llm}^{q_i}), \mathcal{D}_{cg})$ exactly matches to the labeled answer $a_{train}$, we regard $(q_i, \mathcal{Q}_{llm}^{q_i})$ as a question-query pair demonstration.
% We repeat this step until $k$ demonstrations are collected else supplemented from the initial demonstrations.
We repeat this step until $k$ demonstrations are collected. If 
$k$ demonstrations are not collected, we supplement the remaining ones by the initial demonstrations.
\section{Experiments}
\label{sec:experiment}
% \vspace{-3mm}
% To evaluate the effectiveness of our {\model}, 
We adopt 5 datasets covering 3 data types: WikiSQL \cite{DBLP:journals/corr/abs-1709-00103} and WTQ \cite{wtq} for table, WebQuestionsSP(WebQSP) \cite{DBLP:conf/acl/YihRMCS16} and MetaQA \cite{DBLP:conf/aaai/ZhangDKSS18} for KG, and  CronQuestions \cite{DBLP:conf/acl/SaxenaCT20} for temporal KG. Their statistics are shown in Table \ref{tab:datasets} in the Appendix \ref{sec:datasets}. 
% in which we only show the number of QA pairs in the test set not the train and valid set, since our solution is training-free.
% For each dataset, we report the accuracy. Specifically, we compare the answer from NeSyGPT $A_{NeSyGPT}$ to the standard answer $A_{standard}$. If $A_{NeSyGPT} = A_{standard}$, the answer is correct. Otherwise, it is false.
% In the experiments, 

We use GPT-3.5 (gpt-3.5-turbo-0613) as the LLM with self-consistency strategy of 5 times, and SentenceBERT \cite{DBLP:conf/emnlp/ReimersG19} as the dense text encoder.
% for the node-value mapping and retriever.  
% and GPT-4-turbo (gpt-4-1106-preview) under few-shot settings
% with the self-consistency strategy on 5 times and 
% use text-embedding-ada-002 for aligned query. 
If the answer is ``None'' due to mismatched entity-relation pairs and key-value inconsistencies etc., 
% Due to potential issues during query execution, such as mismatched entity-relation pairs and key-value inconsistencies leading to unsuccessful queries resulting in ``None”, 
we implement the retry mechanism with 3 times trials.
% and set the retry limit to 3 for each time, 
% with a prioritized focus on queries that yield executable queries. 
% In the following experiments, 
We set the number of retrieves $m=15$ and the number of demonstrations $k=8$.
% in dynamic demonstration retrieval.
% We also provide statistics on the usage rate and effectiveness of the LLM function in Appendix~\ref{sec:llm_function}.
% The details of demonstrations and metrics for each dataset will be discussed in their respective sections.
% Next, we will introduce the details of experiments on each type of structured knowledge.
% In this section, 
% Next, we introduce the experiments on 3 types of structured data independently from Section \ref{sec:tableqa} - \ref{sec:tkgqa}, and model analysis in Section \ref{sec:modelanalysis}. 
% % considering the datasets, baseline methods, experiment settings are different. 
% % Then in Section \ref{sec:modelanalysis}, we present the model analysis. 
% Finally in Section \ref{sec:potential}, we show the potential of {\model} for two general structured QA tasks with empirical experiments. 

% \subsection{Experiment settings.}
% % 调用的GPT的什么api，每个数据集的demo各多少个，构建demo的原则，使用self consistency的细节，等等
% @jinlong

% ==============================

\subsection{Table QA Experiment}
\label{sec:tableqa}
% \vspace{-2mm}
% \subsubsection{Datasets and Experiment Setting}
% \paragraph{Datasets}
% For fixed-schema knowledge QA, 
% We choose 2 table QA benchmarks, WikiSQL \cite{DBLP:journals/corr/abs-1709-00103} and WTQ \cite{wtq}, where WikiSQL is a large table QA dataset containing over 15 thousand test QA pairs, and WTQ demands more complex reasoning specifically related to tables.
% for example querying content about the previous row in the table. 
\paragraph{Experiment Setting}
 We adopt denotation accuracy \cite{DBLP:conf/emnlp/JiangZDYZW23} to assess whether the predicted answer matches the labeled answer based on set-level equivalence. We write 8 initial demonstrations for both WTQ and WikiSQL. 
 % Additionally, 
 We add the table's column names and one randomly selected record under each column in a linearized format behind the question. 
 % The structure is: \textit{column name: random data $|$ column name: random data ...}
 The initial demonstrations and prompts are shown in Appendix~\ref{sec:wikisql_prompt} and Appendix~\ref{sec:wtq_prompt}. 

% \subsubsection{Result Analysis}

\paragraph{Result Analysis}
As shown in Table \ref{tab:tableQA},
% On WikiSQL, {\model} achieved 85.9\% accuracy under GPT-3.5, surpassing StructGPT, as well as some pre-training methods like MAPO and TAPAS. 
on WikiSQL, compared to models specifically developed for table QA, {\model} surpasses MAPO and TAPAS and approaches UnifiedSKG with an accuracy of 85.9\%.
% demonstrating the significant potential of a trustful framework in table QA. 
Compared to unified models, it achieves a nearly 20\% improvement.
% indicating that the Two-layer Function-based CG Query unleash the generative capabilities of the LLM
% of LLMs by designing $Q_{llm}$ that are easy for LLMs to understand, while 
% and  the translator improves accuracy.

On WTQ, {\model} achieves an accuracy of 44.2\%, surpassing MAPO. However, there remains a gap compared to the other methods. 
We analyze the cases and find three reasons. 
% Firstly, the inherent challenge lies in the question itself. 
Firstly, some questions have more than one correct answer, like \textit{``name a player that had more than 5 league goals but no other goals"}, but only one is included in the labeled answer. 
% Additionally, the emphasis on the subject in natural language may not be reflected in the corresponding table. For example, the question \textit{``total number of countries represented in round 8"} assumes the table containing data only of \textit{round 8} but LLM cannot know this only from the question.
% Some questions include the caption of the table, for example  \textit{``total number of countries represented in round 8"}. However, {\model} doesn't know the topic of the table and tend to think data from the other rounds are also included which causes incorrect LLM queries. 
% Secondly, issues arise because not provide the whole table. Consider questions related to table space information, such as ``how many consecutive wins did the Hoyas have this season?" It requires the whole data of the table to determine the consecutive. There are also challenges related to the specificity of the table, like ``how many countries won a gold medal" where the last row of the table is total data statistics which could make mistakes for the answer.
Secondly, 
some tables are presented in a none standard format,
% issues arise because of non-standard table representation, 
such as the table for \textit{``how many countries won a gold medal"} whose last row is the total statistics of the previous rows. 
Thirdly,  
there are limitations of {\model} during element mapping.
% \textit{Mapping Error}: 
For the question \textit{``What are the number of times a race was held in August?"} the term ``August" could refer to either ``August 1st" or ``August 4th", but the translator only retrieves one instance.
% \textit{Understanding Complex Data}: For instance, determining the winning team under the score of 2-10.
% and not knowing this beforehand can lead to the incorrect answer.
% Thirdly, the question requires complex reasoning out of the capability of {\model}, such as \textit{``how many consecutive wins did the Hoyas have this season?"}. 
% Thirdly, limitations in query design. Instances where the entity to be retrieved in the table corresponds to more than one entity pose challenges. For example, in the question ``what are the number of times a race was held in August," where ``August" could refer to either ``August 1st" or ``August 4th”, the current design returns the closest match as the final result. Additionally, individual table cells may hold rich information, where the ``1-10" under the score column represents two team's scores. In our design, we cannot get the score difference without further decomposition.
\begin{table}[]
% \begin{wraptable}{r}{0.33\textwidth}
    \centering
    % \scriptsize
    \small 
    % \vspace{-12mm}
    \caption{Denotation accuracy of Table QA.}
    \vspace{-2mm}
    \resizebox{.35\textwidth}{!}{
    \begin{tabular}{l|c c}
    \toprule
        \textbf{Methods} & \textbf{WikiSQL} & \textbf{WTQ} \\
        \midrule
        \multicolumn{3}{c}{Data Type Specific Models} \\
        \midrule
        MAPO \cite{DBLP:conf/nips/LiangNBLL18} & 72.6 & 43.8  \\
        TAPAS \cite{DBLP:conf/acl/HerzigNMPE20}  & 83.6 & 48.8 \\
        UnifiedSKG \cite{DBLP:conf/emnlp/XieW0ZSYWZYWZWL22} & 86.0  &    49.3 \\
        TAPEX \cite{DBLP:conf/iclr/LiuCGZLCL22} & \textbf{89.5}  & 57.5 \\
        DATER \cite{DBLP:conf/sigir/YeHYLHL23}	&- & \textbf{65.9}	 \\
        \midrule
        \multicolumn{3}{c}{Unified Models (with GPT3.5 as the LLM)} \\
        \midrule
% StructGPT(Davinci-003) & 49.1 	& 34.8	 \\
% - IRR	& 51.8 & 39.2	 \\
% - IRR (few-shot)	& 64.6  & 57.0	\\
% StructGPT(ChatGPT)	& 51.6  & 43.3	\\
% - IRR	& 54.4 & 48.4	 \\
% - IRR (few-shot)	& 65.6  & 52.2	\\
StructGPT \cite{DBLP:conf/emnlp/JiangZDYZW23} & 65.6  & 52.2	\\
Readi \cite{cheng2024call} & 66.2 & \textbf{66.7} \\
% \midrule
% \model(GPT3)	& 84.5 & 46.1	 \\
\textbf{\model}(ours)	& \textbf{85.9} & 44.2	 \\
% \model(GPT4)  & 89.0 & 49.8  \\
         \bottomrule
    \end{tabular}
    }
    % \vspace{-10mm}
    \label{tab:tableQA}
% \end{wraptable}
\end{table}

% \vspace{-2mm}
% ==============================
\subsection{Knowledge Graph QA}
% \vspace{-2mm}
\begin{table}[]
% \begin{wraptable}{r}{0.5\textwidth}
    \centering
    % \scriptsize
    % \vspace{-10mm}
    \caption{Hit@1 results of KG QA. Digits with $^*$ are set comparison accuracy.  
    % Results with $^*$ is regarding {\model}'s answer is correct if the gold answer is a subset of it.
    }
    \vspace{-2mm}
        \resizebox{.45\textwidth}{!}{
    \begin{tabular}{ l | c c c | c }
    \toprule
    \textbf{Methods} & \multicolumn{3}{c}{\textbf{MetaQA}}  & \textbf{WebQSP} \\
    & \textit{1 hop} & \textit{2 hops} & \textit{3 hops} & \\
    \midrule
        \multicolumn{5}{c}{Data Type Specific Models} \\
    \midrule 
    KV-Mem 
    \cite{DBLP:conf/emnlp/MillerFDKBW16}
    & 96.2	& 82.7	& 48.9 & 46.7\\
GraftNet 
\cite{DBLP:conf/emnlp/SunDZMSC18}	
& 97.0	& 94.8	& 77.7 & 66.4 \\
EmbedKGQA 
% \cite{DBLP:conf/acl/SaxenaTT20}
\shortcite{DBLP:conf/acl/SaxenaTT20}
& \textbf{97.5}	& 98.8	& 94.8 & 66.6 \\
NSM 
\cite{DBLP:conf/wsdm/HeL0ZW21}	
& 97.1	& \textbf{99.9}	& 98.9 	& 68.7 \\
UniKGQA 
\cite{DBLP:conf/iclr/JiangZ0W23}
& \textbf{97.5}	& 99.0	& 99.1 &  \textbf{75.1} \\
% \midrule 
Davinvi-003 
\cite{DBLP:conf/nips/Ouyang0JAWMZASR22} 
& 52.1 & 25.3 & 42.5 & 48.3 \\
KB-BINDER 
\cite{DBLP:journals/corr/abs-2305-01750} 
& 93.5 & 99.6 & 96.4 & -\\
KB-BINDER-R 
\cite{DBLP:journals/corr/abs-2305-01750} 
& 92.9 & \textbf{99.9} &  \textbf{99.5} & -\\
\midrule
        \multicolumn{5}{c}{Unified Models (with GPT3.5 as the LLM)} \\
        \midrule 
% StructGPT(Davinci-003)		& 52.1	& 25.3	& 42.5 & 48.3 \\
% - IRR \cite{DBLP:conf/emnlp/JiangZDYZW23}		& 94.4	& 59.5	& 70.2 & 71.9 \\
% - IRR (few-shot)	& 97.1	& 93.5	& 75.3 & 71.0	 \\
% StructGPT(ChatGPT)		& 61.9	& 31.0	& 43.2 & 61.2 \\
% - IRR		& 94.2	& 93.9	& 80.2 & 72.6 \\
% - IRR (few-shot)		& 97.1	& 97.3	& 87.0 & 69.6\\
StructGPT 
\cite{DBLP:conf/emnlp/JiangZDYZW23}
& 97.1	& 97.3	& 87.0 & 69.6\\
Readi 
\cite{cheng2024call} 
& \textbf{98.4} & \textbf{99.9} & \textbf{99.4} & 74.3 \\ 
% \midrule 
% \model(GPT3.5)  & 97.3	& 98.4	& 98.6  & 70.9 \\
%  & \textbf{99.9}*	&  {99.8}*	& \textbf{99.9}* & 66.1$^{\$u}$/89.0$^{\$c}$\\
 % \model(GPT3.5)  & {97.1}	& {97.9}	& {98.4}  & {81.8} \\
 % & \textbf{99.94}*	&  \textbf{99.99}*	& \textbf{99.9}* & 79.1$^{\$u}$/90.6$^{\$c}$\\
 \textbf{\model}(ours) & 97.1 & 97.9 & 98.4 &  \textbf{83.5} \\
 
 & 99.94$^*$	&  99.99$^*$	& 99.99$^*$ &  \\
% NeSyGPT(GPT4) & -  & - & - & \textbf{76.5} \\
%  & -  & - & - & 73.0$^{\$u}$/89.5$^{\$c}$ \\
\bottomrule
    \end{tabular}
    }
    \vspace{-3mm}
    
    \label{tab:kgqa}
    % \vspace{-3mm}
    % \end{wraptable}
\end{table}
% \subsubsection{Datasets and Experiment Setting}
% \paragraph{Datasets} We choose 2 KGQA benchmarks, WebQuestionsSP(WebQSP) \cite{DBLP:conf/acl/YihRMCS16} and MetaQA \cite{DBLP:conf/aaai/ZhangDKSS18}, which are constructed from Freebase and Wikidata. MetaQA includes 1-3 hops questions, thus questions included in MetaQA could be classified into  1 hop, 2 hops, and 3 hops. 

\paragraph{Experiment Setting}
% Following existing works, we report the Hit@1 results. 
% While {\model} only being able to get all answers that satisfy the queries, we choose one answer (the first one) as the top-1 for validation. 
% If a list of answers are given by {\model}, we choose the first one as the top-1 for validation.

% MetaQA has 9 relations and has templates for 2-hop and 3-hop.
For MetaQA, we construct 13, 15, and 11 demonstrations for 1-hop, 2-hop, and 3-hop, respectively. The instruction of the prompt includes a description of CG data and a list of relations for the LLM to choose when generating queries. 
% We mask the topic entity in the question and use the remaining part of the question as a template for dynamic demonstration retrieval. 
% In subsequent experimental analyses, we further scrutinize the accuracy of all generated answers. 
% In the following experimental analyses, we evaluate the accuracy of all generated answers.
The initial demonstrations and prompts are shown in Appendix \ref{sec:meta_prompt}.

For WebQSP, we used a processed version of the official WebQSP Dataset release\footnote{https://nlp.stanford.edu/projects/kgqa/webqsp.zip}. For each question, a relevant subset of the KG (a set of triples) is retrieved from Freebase. Some relations and entities in WebQSP represented by IDs without entity names lack clear semantic meaning. 
% To explore {\model}'s performance on semantically explicit questions, 
Thus we divide the questions in WebQSP into two types, questions regarding semantic-clear and semantic-unclear relations which can be found in Appendix \ref{sec:wqsp_detail}. 
% A relation is a semantic-clear if 
% there are head entity and tail entity in a triple of it have names in natural language, otherwise, it is a semantic-unclear relation.
% A relation is regarded as semantic-clear if there are head entity and the tail entity in its triples have names in natural language. 
The initial demonstrations include topic entity, first step relations (relations have topic entity) and second step relations for WebQSP. More details are shown in Appendix \ref{sec:wqsp_prompt}.

% The format of provided relations differs between the two types of categories. Specifically, for questions with semantic-clear relations, the provided relations include the head entity, relation, and tail entity, expressed as $(h, r, t)$. But for questions of semantic-unclear relations, 
% % where corresponding head and tail entities with semantic meaning cannot be found, 
% only the relations are listed without explicit entities.

% \begin{wrapfigure}{l}{0.38\textwidth}
% \begin{figure}
%     \centering
%     \includegraphics[width=0.3\textwidth]{figures/metaqa_number.pdf}
%     \caption{The frequency of numbers of  answers labeled and answers predicted by our method on MetaQA.}
%     \label{fig:number_metaqa}
% \end{figure}

\begin{table}[]
% \begin{wraptable}{r}{0.5\textwidth}
    \centering
    % \scriptsize
    \vspace{-4mm}
    \caption{Hits@1 of temporal KG QA on CronQuestion.}
    \vspace{-2mm}
    \resizebox{.45\textwidth}{!}{
    \begin{tabular}{l |c| c c |c c  }
    \toprule
    \textbf{Methods} & & \multicolumn{2}{c}{ \textbf{Question Type}} & \multicolumn{2}{c}{\textbf{Answer Type}} \\
    % & &Complex & Simple & Entity & Time \\
     &\textit{All} &\textit{Com} & \textit{Sim} & \textit{Ent} & \textit{Tim} \\
    \midrule
    \multicolumn{6}{c}{Data Type Specific Models} \\
    \midrule
BERT  \cite{DBLP:conf/naacl/DevlinCLT19}	 & 24.3	& 23.9	& 24.9	& 27.7	& 17.9 \\
RoBERTa  \cite{DBLP:journals/corr/abs-1907-11692}	 & 22.5	 & 21.7	 & 23.7	 & 25.1	 & 17.7 \\
EmbedKGQA  
% \cite{DBLP:conf/acl/SaxenaTT20}
\shortcite{DBLP:conf/acl/SaxenaTT20}
& 28.8	& 28.6	& 29.0	& 41.1	& 05.7 \\
EaE  \cite{DBLP:conf/emnlp/FevrySFCK20}	& 28.8	& 25.7	& 32.9	& 31.8	& 23.1 \\
CronKGQA  \shortcite{DBLP:conf/acl/SaxenaCT20}	& 64.7	& 39.2	& 98.7	& 69.9	& 54.9 \\
% EntityQR	& 0.745	& 0.562	& 0.990	& 0.831	& 0.585 \\
TempoQR-Soft  \shortcite{DBLP:conf/aaai/MavromatisSIAHG22}	& 79.9	& 65.5	& 99.0	& 87.6	& 65.3 \\
TempoQR-Hard  \shortcite{DBLP:conf/aaai/MavromatisSIAHG22}	& 91.8	& 86.4	& 99.0	& 92.6	& 90.3 \\
TSQA  \cite{DBLP:conf/acl/ShangW0022}	& 83.1	& 71.3	& 98.7	& 82.9	& 83.6 \\
TMA  \cite{DBLP:conf/icassp/LiuLFWWJ23}	& 78.4	& 63.2	& 98.7	& 79.2	& 74.3 \\
CTRN  \cite{DBLP:journals/apin/JiaoZWZQWZL23}	& 92.0	& 86.9	& 99.0	& 92.1	& 91.7 \\
LGQA  \cite{DBLP:conf/ijcai/LiuLLGLWW0FG23}	& {96.9}	& {94.5}	& {99.2}	& {96.2}	& 96.6  \\
\midrule 
\multicolumn{6}{c}{Unified Models (with GPT3.5 as the LLM)} \\
\midrule
% \model (GPT3) &0.960 &0.935 & \textbf{0.994}	&0.954 & \textbf{0.972}  \\
% \model (GPT4) &\underline{0.961} &\underline{0.938} & 0.991	&\underline{0.956} & \underline{0.971}   \\
\textbf{\model}(ours) & 
\textbf{97.2} & \textbf{95.4} & \textbf{99.5} & \textbf{96.1} & \textbf{99.1} \\
\bottomrule
    \end{tabular}
        }
\vspace{-3mm}
    \label{tab:temporalQA}
    % \vspace{-4mm}
    % \end{wraptable}
\end{table}

% \begin{table}[]
%     \centering
%     \begin{tabular}{c|c}
%          &  \\
%          & 
%     \end{tabular}
%     \caption{Caption}
%     \label{tab:my_label}
% \end{table}

% \vspace{-2mm}
\paragraph{Result Analysis}
% For MetaQA, our use of GPT-3.5 in 1-hop, 2-hop, and 3-hop achieved competitive results under the Hits@1 metric. 
We report the Hit@1 results in Table \ref{tab:kgqa}.

For MetaQA, {\model} performs competitively
in comparison with existing unified methods. 
% But during experiments, we find the name ambiguity in MetaQA (e.g., two movies may have the same name, and it is unclear which one is referred to when the name appears in a question), and the gold answer might be a subset of the true answers. 
% During our experiments, we discovered name ambiguity issues in MetaQA. For instance, two movies might share the same name, making it unclear which one is referenced in a question. Therefore, the gold answer might only represent a subset of the true answers.
% Based on this analysis, we think the higher hit@1 results might not means better QA performance on MetaQA, especially the hit@1 results are already larger than $97\%$.
% Thus we report the accuracy of set comparison between predicted answers and labeled answers (the last row in Table \ref{tab:kgqa}), where we regard the predicted answers is correct it includes all labeled answers. {\model} achieves above $99.9\%$ accuracy on all the three types of questions. 
During our experiments, we discover name ambiguity issues in MetaQA. For instance, two movies might share the same name, making it unclear which one is referenced in a question. Consequently, the gold answer might only represent a subset of the true answers.
Based on this analysis, higher Hit@1 results may not indicate better QA performance on MetaQA, especially when the Hit@1 results exceed 97\%, due to the incompleteness of labeled answers. Therefore, we report the accuracy of set comparison between predicted and labeled answers (last row in Table \ref{tab:kgqa}), where predicted answers are considered correct if they include all labeled answers. {\model} achieves over 99.9\% accuracy across all three types of questions.
% But {\model} gives all the correct answers $A_{{\model}}$ based on the correct query. Thus if the labeled answer $A_{standard} \subseteq A_{{\model}}$, we regard our answer as correct. 
% Results are shown in Table \ref{tab:kgqa}. {\model} achieves above $99.9\%$ accuracy on all the three types of questions. 
% In this case, 
% our experimental results, indicated in the following line, 
% demonstrate achievement of 
% 99.9\%, 99.8\%, and 99.9\% on the three datasets respectively. 
% In Figure~\ref{fig:number_metaqa}, we 
Figure \ref{fig:combined}(a)
shows the frequency of numbers of answers predicted by {\model} and answers labeled.
% We count the number of elements under label and prediction at the same time in Figure~\ref{fig:number_metaqa}, and 
It shows the two frequency distributions are nearly identical, and thus we can conclude the
{\model} performs perfectly on MetaQA and this is not due to giving a large number of answers. 
% mitigating the rise in Hit@1 resulting from excessively large $A_{UniT}$. 
% This demonstrates {\model}'s strong capability in understanding simple semantic questions.

For WebQSP, {\model} achieves a Hit@1 score of $83.5\%$,  surpassing all the baselines and achieves the state-of-the-art. 
% More detail can be found in Appendix \ref{sec:wqsp_detail}.
% Furthermore, we achieve a state-of-the-art accuracy of 76.5\% on GPT4.
% Notably, we observe a significant difference between the results of questions with semantic-clear and -unclear relations, which are $91.0\%$  and $81.5\%$ respectively. Thus proper relation and entity naming in the question is important for understanding its meaning by the LLM.
% Specifically, the accuracy for questions with semantic-clear relations is 90.6\% 
% is higher by 23\% for GPT-3.5 and 16\% for GPT-4 compared to semantic-unclear ones. Additionally, for semantic-clear cases, GPT-4 
% % shows an improvement of nearly 7\%
% gains only 0.5\% on Hit@1 compared to GPT-3.5. 
% This is because the accuracy of answering questions with semantic-clear relations is close to the upper accuracy of retrieved WebQSP. 

% The main reason is that the accuracy of clear semantic problems is close to the upper accuracy of retrieved WebQSP, so there is little room for improvement but huge for the unclear.
% This improvement is attributed to GPT-4's enhanced semantic understanding, coupled with the fact that questions with unclear semantic relations have more room for improvement. On the other hand, the improvement is less pronounced for questions with clear semantic meaning, as GPT-3.5 already achieves high accuracy in such cases, leaving less room for enhancement.

% \input{tables/ErrorType}

% ==============================
\subsection{Temporal Knowledge Graph QA}
\label{sec:tkgqa}
% \vspace{-2mm}
% \subsubsection{Datasets and Experiment Setting}
% \paragraph{Datasets}
% For temporal knowledge QA, we choose a recently widely used dataset CronQuestions  \cite{DBLP:conf/acl/SaxenaCT20}, which is constructed from temporal knowledge graph in Wikidata. According to the question type, the questions are classified into complex and simple questions. According to the answer type, the questions are classified into entity questions and time questions. 

\paragraph{Experiment Setting}

We use Hits@1 for evaluation.%whether the top-1 predicted answer is correct. 
%Similarly, we select the first predicted answer from the adaptive results for validation as top-1. 
 % In the subsequent experiments, 
 We design 9 demonstrations, each including the question, relation, and annotation. 
 % Additionally, 
 Entities that are not originally presented in the annotation but exist in the entity list are appended at the end. We use the question and answer type of each question as metadata during retrieving. The initial demonstrations and prompts are shown in Appendix \ref{sec:cron_prompt}.

\paragraph{Result Analysis}
% On CronQuestions, 
As Table \ref{tab:temporalQA} shows,
{\model} is the only unified model for this task and achieves the state-of-the-art.
% The performance of {\model} using GPT-3.5 and GPT-4 are close, which is different from what is observed on other datasets. 
% This may be because of the mismatching between the gold answer and the question.
% This may be because the unclear logic in query is frequently with both of them. 
% Some questions in the dataset are natural language ambiguous, 
% During the experiments, some questions are annotated by negative gold answers.
% For example, for the question  ``Who was the president before \textit{Gaspard}?" where \textit{Gaspard} was president from 1857-1868,  \textit{Alfred} who was the president for 1852-1870 is included in the gold answer but \textit{Charles} who was the president for 1848-1869 is not. Both \textit{Alfred} and \textit{Charles} should be the answer. 
% Thus the quality of the datasets should be improved. 
Since {\model} and LGQA could achieve high performance on CronQuestion, more challenging QA datasets about temporal KG are expected to be created. 
% However, this is not a problem for model trained on the training data, since there are patterns of constructing the QA pair, so the trained model can learn the pattern and make prediction equal to the gold answer even though the answer does not match the semantic meaning of the question. 

% But the semantic is important for Natural 
% But the mismatch problem can not be solved by improving the NL understanding capability of the LLM in our solution. 
% but it is difficult to understand them from the perspective of natural language. 
% Even so, our method achieves sota in simple and time types, and the overall is comparable to sota. 

\subsection{Model Analysis}
\begin{figure*}
    \centering
        \vspace{-4mm}
    \includegraphics[width=\textwidth]{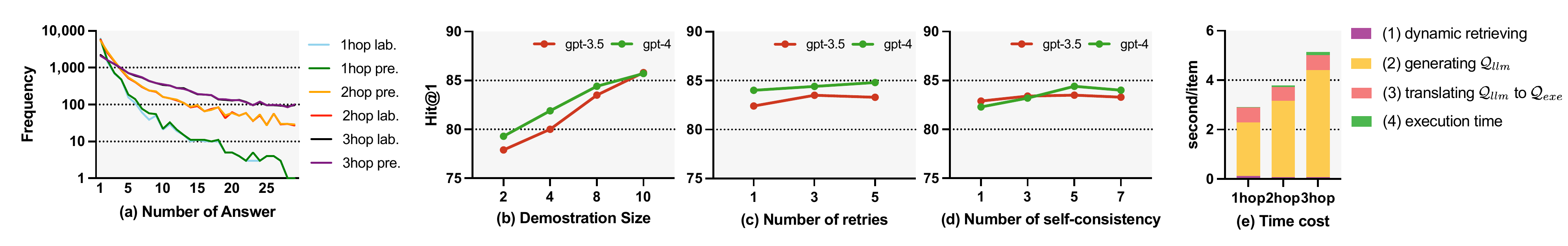}
        \vspace{-6mm}
    \caption{(a) The frequency of numbers of answers labeled and predicted by ours on MetaQA. (b) - (d) The hyperparameter of demostration size, retry and self-consistency on WebQSP. (e) The time cost of MetaQA.}
    \vspace{-2mm}
    \label{fig:combined}
\end{figure*}

\label{sec:modelanalysis}
\begin{figure}[]
    \centering
\includegraphics[width=0.5\textwidth]{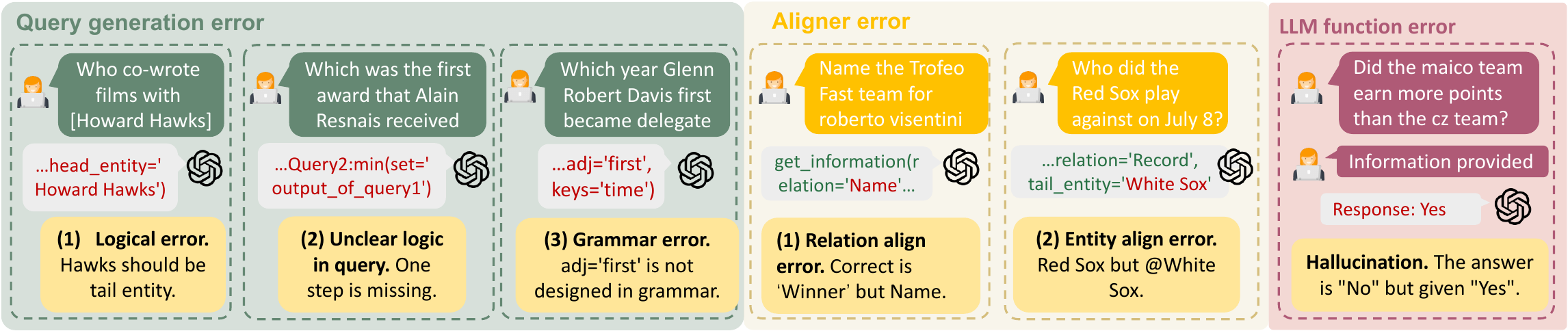}
\vspace{-5mm}
    \caption{Error cases of {\model}.}
    \label{fig:error_case}
    \vspace{-3mm}
\end{figure}

\paragraph{Ablation Study}
\begin{table}[]
    \centering
        \vspace{-4mm}
    \caption{Ablation Study. 
    % Denotation accuracy of WikiSQL and WTQ, Hit@1 of the MetaQA, WebQSP and CronQuestion. 
    \textit{-dynamic demo} means using static initial demonstration. 
    \textit{-two-layer query} means making LLM directly generate the execution query.
    The result of MetaQA is the average of 1,2,3 hops.}
    \vspace{-2mm}
    \resizebox{.45\textwidth}{!}{
    \begin{tabular}{l|c c c c c }
    \toprule
         & \textbf{WikiSQL} & \textbf{WTQ} & \textbf{MetaQA} & \textbf{WebQSP} & \textbf{CronQuestion} \\
    \midrule
        TrustUQA & \textbf{85.9} & \textbf{44.2}  
        % & 97.1/99.94* & 97.9/99.99 & 98.4/99.99 
        & 97.8/\textbf{99.97*}
        & \textbf{83.5} & \textbf{97.2}  \\
        \textit{-dynamic demo} & 84.5 & 43.7 
        % & 97.3/99.94 &98.4/99.8 & 98.6/99.9 
        & \textbf{98.1}/99.87* 
        & 75.1 & 96.1  \\
        \textit{-two-layer query} & 84.4 & \textbf{44.2} 
        & 97.7/99.87* 
        & 83.4 & 87.6 \\
        \bottomrule
    \end{tabular}
    }
    
    \label{tab:ablation}
\end{table}
As shown in Table ~\ref{tab:ablation}, without dynamic demonstration retrieval (\textit{-dynamic demo}) or without two-layer query functions (\textit{-two-layer query}), the performance of {\model} decrease in majority of the datasets.
% exhibits subpar performance campared with ours in 5 structured data consistently, which proves the effectiveness of dynamic demo retrieving. And without two-layer, experiments show the Decreased performance especially in CronQuestion which certifies $\mathcal{Q}_{llm}$ function is designed to allow better understanding in complex knowledge representation and reasoning.

\paragraph{Using Different LLMs and Text Encoders}
% \begin{table}[]
%     \centering
%     \resizebox{0.3\textwidth}{!}{
%     \begin{tabular}{l l| c c }
%     \toprule
%        LLM  & Text Encoder & WikiSQL & WTQ \\
%        \midrule
%        \multirow{4}{*}{GPT-3.5} & SentenceBert & 85.9 & 44.2 \\ 
%        & DPR & 86.0 & 44.5 \\
%        & ANCE & 85.5 & 45.3 \\
%        & M3E & 86.1 & 44.5 \\
%        \midrule
%        \multirow{4}{*}{GPT-4} & SentenceBert & 87.4 & 52.1 \\
%        & DPR & 87.4 & 51.4 \\
%        & ANCE & 87.4 & 51.2 \\
%        & M3E & 87.7 & 50.8 \\
%        \bottomrule 
%     \end{tabular}
%     }
%     \caption{{\model} with different LLMs and text encoders. }
%     \label{tab:my_label}
% \end{table}

\begin{table}[]
    \centering
    \caption{{\model} with different LLMs and text encoders. }
    \vspace{-2mm}
    \resizebox{0.45\textwidth}{!}{
    \begin{tabular}{l|c c c c | c c c c}
    \toprule
         & \multicolumn{4}{c|}{GPT-3.5} & \multicolumn{4}{c}{GPT-4}  \\
         % \cline{2-9}
         & SentB & DPR & ANCE & M3E & SentB & DPR & ANCE & M3E\\
         \midrule 
         WikiSQL & 85.9 & 86.0 & 85.5 & 86.1 & 87.4 & 87.4 & 87.4 & 87.7  \\
         WTQ & 44.2 & 44.5 & 45.3 & 44.5  & 52.1 & 51.4  & 51.2 & 50.8 \\
         \bottomrule
    \end{tabular}
    }
    \vspace{-2mm}
    % \caption{Caption}
    \label{tab:vary_llm_encoder}
\end{table}
We have tested GPT-4 and other three text encoders on two Table QA datasets. The results can be seen in Table \ref{tab:vary_llm_encoder}. The results show that TrustUQA performs slightly differently with various text encoders, and consistently achieves better performance with the more advanced LLM GPT-4 compared to GPT-3.5.

\paragraph{Efficiency Analysis}
We evaluate the time cost of each step of our method on the MetaQA dataset with 100 random samples\footnote{Our system is equipped with 2*NVIDIA A100 PCIe 40GB GPUs, 40 physical cores across 2 sockets, each socket containing 20 cores. The Intel Xeon Gold 6148 processors operate at a base speed of 2.40 GHz, with a maximum of 3.70 GHz.}. 
Steps include (1) \textit{dynamic demo retrieval}, (2) \textit{generating $\mathcal{Q}_{llm}$}, 
% (3) mapping variables from $\mathcal{Q}_{llm}$ to nodes in the CG data ($T_Q(mapping)$), (4) translating $\mathcal{Q}_{llm'}$ to $\mathcal{Q}_{exe}$ following translation rules ($T_Q(rule)$), 
(3) translating $\mathcal{Q}_{llm}$ to $\mathcal{Q}_{exe}$ 
and (4) execute the query. 
Results are in Figure \ref{fig:combined}(e).
We can see that most of the time are cost for $\mathcal{Q}_{llm}$ generation. The dynamic demonstration retrieval, query translation and execution are quite efficient which takes less than 20\% of the whole time. 

% There are two steps related to the two-layer CG query function design, as illustrated in Translation paragraph in the 3.2 Query Translator: (1) mapping variables from $\mathcal{Q}_{llm}$ to nodes in the CG data ($T_Q(mapping)$), and (2) translating $\mathcal{Q}_{llm'}$ to $\mathcal{Q}_{exe}$ following translation rules ($T_Q(rule)$). Among these, mapping variables is a necessary step regardless of the query function design, as the variables need to be mapped to nodes to make the query executable. Therefore, the step of translating the query according to the rules is the only one specifically added by the two-layer CG query function design. From the figure, this additional translation step only accounts for 2\%-13\% of the total time required to get the answer. Thus, the two-layer CG query function not only effectively manages complexity but also maintains efficiency during application.

\paragraph{Hyperparameter Analysis}
In Figure \ref{fig:combined} (b)-(d), we show that more demonstrations and more re-try leads to slightly better results. And generating answers in the self-consistency strategy by 5 times achieves the best results. Considering the tradeoff between time/computation cost and the performance, we set a moderate number for demonstrations and retry, which are 8 and 3 respectively.

\paragraph{Error Analysis}
\label{sec:erroranalysis}
% We categorize the error cases of {\model} into three types.
We recognized three types of errors, cases shown in Figure \ref{fig:error_case}.
% \textit{query generation error}, \textit{Aligner error}, and \textit{LLM function error}. 
% Figure \ref{fig:error_case} shows several cases.
(1) The most frequent type is \textit{query generation error}, such as logical error where the generated query is logically wrong, unclear logic in query with steps either missing or unnecessary, and  grammar error in query.
% . 1) \textbf{Logical errors} in queries, such as confusion between head and tail entities. For instance, in the question ``Who co-wrote films with [Howard Hawks]," when the relation ``written\_by" is chosen, Howard Hawks should be the tail entity, but in the first query generation step, it is incorrectly identified as the head entity. 2) \textbf{Unclear logic in execution} steps leading to too many or too few steps. For example, in the question ``When Nuno Espírito Santo played the last game," the system should retrieve the specific time instead of continuing to query what the last game is. Similarly, in ``Which was the first award that Alain Resnais received," the system should continue querying for the award after obtaining the time instead of stopping. 3) \textbf{Queries not following the defined grammar}, like in ``Which year Glenn Robert Davis first became delegate," where the generated get\_information function includes adj=`first,' which was not specified in our function design. 4) \textbf{Errors in entity and relation generation}, as seen in ``Which purpose has an Elevation + Height of 0 + meters (ft), a Location of bikini, namu (charlie), and a Yield of 220 kt?" where the generated get\_information function incorrectly assigns `220' as the tail entity instead of `220 kt.' In the question ``What are fun things to do in Sacramento, CA?" the system wrongly identifies the relationship to solve the problem as `location.location.contains' instead of the correct `travel.tourist\_attraction.near.'
(2) The second type is \textit{mapping error} including errors in relation and entity mapping during the parameter-value mapping step respectively.
% \textbf{1) Errors in relation mapping}, as in the question ``Name the Trofeo Fast team for Roberto Visentini," where the retrieved relation should be `winner' instead of `name.' \textbf{2) Errors in entity mapping}, for example, in ``Who did the Red Sox play against on July 8?" where the retrieved entity should be `Red Sox,' but it is incorrectly `@ White Sox.'
(3) The third type is \textit{LLM function error}. When calling LLM function, though enough data are provided, sometimes LLM still generates wrong answers.
% For instance, in the question ``Did the Maico team earn more points than the CZ team?" after obtaining the scores, the LLM function response incorrectly states ``Yes," which contradicts the provided data.

% \paragraph{Frequency of calling LLM function.}
\paragraph{Calling Rate and Result of LLM function.}\label{sec:llm_function}

\begin{table}[]
    \centering
    % \vspace{-2mm}
    \caption{The calling rate and results of the LLM function.}
    \vspace{-2mm}
    \resizebox{0.45\textwidth}{!}{
    \begin{tabular}{l|c c c c c}
    \toprule
         & \textbf{WikiSQL} & \textbf{WTQ} & \textbf{MetaQA} & \textbf{WQSP} & \textbf{CronQuestion} \\
         \midrule
        Calling Rate  & 4.0\% & 30.7\% &  $<$1\% & $<$1\% & 2.0\% \\
        Results & 38.9\% & 11.4\% & 0.0\% & 2.2\% & 20.4\% \\
        \bottomrule
    \end{tabular}
    }
    \vspace{-3mm}
    \label{tab:statistics}
\end{table}

% Table \ref{tab:statistics} presents the usage and accuracy of the LLM function on five datasets during testing, Acc. For WikiSQL and WTQ is denotation accuracy while the others are Hit@1. 
% The results refer to denotation accuracy for WikiSQL and WTQ and Hit@1 for the others. 
% As shown in Table \ref{tab:statistics}, 
Results are shown in Table \ref{tab:statistics}.
We observe the following:
% \textit{1) The overall usage of LLM function is low.} 
\textit{(1)
Except for WTQ, the usage rates under each dataset are generally low, with most being less than 1\%. }
% In the case of WTQ, 
% The higher usage of LLM function in WTQ is primarily attributed to its involvement in addressing issues that designed functions find challenging, 
The higher call rate in WTQ is mainly because there are questions the designed functions struggle to address,
such as questions like `who gets more score, A or B?'.
\textit{(2) The overall accuracy is low.} Compared to the overall experimental results on each dataset, the results of the LLM function are consistently lower. 
Thus calling the execution functions we design in {\model} is more trustful. 
% However, due to its relatively small proportion, its impact on the overall experimental results is minor. 
% Simultaneously, considering its frequent usage in WTQ, enhancing performance in this context becomes a significant focus for future improvements.

% Analysis with GPT4 as the LLM can be found in Appendix~\ref{sec:gpt4}. 

% \input{tables/tableQA_gpt4}
% \paragraph{Stronger LLMs} The results with GPT-4 are shown in Table \ref{tab:tableQA_gpt4}.
% In WikiSQL, {\model} outperformed UnifiedSKG and improved by 1.5 over GPT3.5 which comes to 87.4. 
% While in WTQ, the performance further improved, and outperformed UnifiedSKG with an accuracy of 52.1\%.
% This shows {\model}'s proficiency in deep semantic understanding, and indicates that {\model} will perform better as the ability of LLM itself improves.

\subsection{Potentials of {\model}}
\label{sec:potential}
% To further demonstrate the generality of {\model} by question whose answers may rely on different data sources, we experience two tasks: \textbf{Mixed Structured Knowledge QA} and \textbf{Cross Structured Knowledge QA}.
\paragraph{QA over Mixed Structured Data}
% Mixed Structured Knowledge QA where 
In this task,
each question relies on one of the mixed multiple given data sources but which data source it relies on is not specified, which we call \textit{mixed data ($D_{\text{mixed}}$) + answer source unknown ($S_{\text{unknown}}$)} setting. 
%\textit{mixed data} means data from different source are mixed and stored together. 
% In contrary, \textit{unmixed data} setting means we transform data from into CG and stored them independently. This setting will effects the dynamic demonstration retrieving results.
%\textit{answer source unknown} means we don't know which data source is queried given a question. 
% In contrary, \textit{answer source known} means which source is queried. 
% This setting will effects the unified or data-source-specific prompt to use in the experiment.  
In contrast, the setting of the evaluation tasks in Section \ref{sec:tableqa} - \ref{sec:tkgqa} is \textit{unmixed data ($D_{\text{unmix}}$) + answer source known ($S_{\text{known}}$)}. 
% Since there is no corresponding dataset currently, we use the existing structured data set for construction. 
% To simulate QA over mixed structured data with \textit{mixed data + answer source unknown} setting, 
To simulate this scenario, 
we randomly extract 100 data entries from each of WikiSQL, WTQ, MetaQA-1hop, WebQSP, and CronQuestion, and experiment on these 500 test questions. 
During experiments, for $D_{\text{mix}}$ setting, we translate data from 5 sources into CG form and store them together, and we store them independently for $D_{\text{unmix}}$ setting.
For $S_{\text{unknown}}$ setting, we use a unified prompt with the demonstration's format only including the question, and for $S_{\text{known}}$ setting, we use the same prompt for each data source as in the main experiments. 
% % Please add the following required packages to your document preamble:
% % \usepackage{booktabs}
% \begin{table}[]
% \centering
% \caption{Mixed structured Knowledge QA for 5 datasets.}
% \label{tab:mixedQA}
% \begin{tabular}{@{}l|ccccc|c@{}}
% \toprule
% \multicolumn{1}{c|}{Dataset} & WikiSQL & WTQ & \begin{tabular}[c]{@{}c@{}}MetaQA \\ 1-hop\end{tabular} & WebQSP & CronQuestion & Avg. \\ \midrule
% native dataset & 85  & 42  & 99   & 84  & 95  & 81 \\
% \begin{tabular}[c]{@{}l@{}}+w/o structures w/ source\end{tabular}  & 79  & 25  & 97   & 46  & 84  & 66 \\
% \begin{tabular}[c]{@{}l@{}}+w/o structure w/o source\end{tabular} & 56  & 8   & 93   & 37  & 72  & 53 \\
% dynamic demo. Acc & 96\% & 42\% & 100\% & 51\% & 85\% &    \\ \bottomrule
% \end{tabular}
% \end{table}

% \usepackage{booktabs}
% \usepackage{multirow}

\begin{table}[h]
% \begin{wraptable}{r}{0.6\textwidth}
\centering
% \scriptsize
\vspace{-3mm}
\caption{Results of QA over mixed structured data.}
\vspace{-2mm}
\label{tab:mixedQA}
\resizebox{.45\textwidth}{!}{
\begin{tabular}{l|ccccc}
\toprule
 &
  WikiSQL &
  {WTQ} &
  % \begin{tabular}[c]{@{}c@{}}MetaQA \\ 1-hop\end{tabular} 
  {MetaQA}
  &
  {WebQSP} &
  {CronQ.} \\ % & \multicolumn{1}{l}{Avg.} \\ 
  \midrule
% mixed data + answer source unknown
$D_{\text{mix}}$ + $S_{\text{unknown}}$
& 56.0   & 8.0    & 93.0    & 37.0   & 72.0  \\ % & 53 \\
dynamic demo. Acc                                                           & 96\% & 42\% & 100\% & 51\% & 85\% \\ % &    \\ 
\midrule 
% unmixed data + answer source unknown
$D_{\text{unmix}}$ + $S_{\text{unknown}}$
& 79.0   & 25.0   & 97.0    & 46.0   & 84.0  \\ % & 66 \\
% \begin{tabular}[c]{@{}l@{}}\textit{+w/o structure w/ merge}\end{tabular} 
% unmixed data + answer source known  
$D_{\text{unmix}}$ + $S_{\text{known}}$
& 85.0   & 42.0   & 99.0    & 84.0   & 95.0   \\ % & 81 \\
% \begin{tabular}[c]{@{}l@{}}\textit{+w/o structure w/o merge}\end{tabular}  

\bottomrule
\end{tabular}
}
% \end{wraptable}
\end{table}

Results are shown in Table~\ref{tab:mixedQA}, in which we report the results using the same evaluation metric on questions from each dataset. 
The results in the first-row show that {\model} could answer questions for different data sources and thus has the potential for QA over mixed structured data. But compared to the original results, i.e. the $D_{\text{unmix}} + D_{\text{known}}$ 
% \textit{unmixed data + answer source known} 
setting, performance drop occurs because the number of candidates for node-value mapping is significantly larger.
% the mixed data caused errors in node-value mapping. 
Comparing the last two rows, we observe a significant performance drop. This is because the results are positively correlated with the dynamic demonstration retrieval accuracy, whether the dynamic demonstrations are retrieved from the correct data source, as shown in the second row in Table~\ref{tab:mixedQA}. We refer the reader to Appendix~\ref{exp:mixed} for more details of this experiment. 
% which could answer questions  different  good performance on MetaQA questions. 
% But compared to the 

% From the perspective of data, there are two settings, \textit{mixed data} with all data from different source mixed together, and \textit{unmixed data} with all data from different source stored independently. The data setting will effects the dynamic demonstration retrieving results.  
% From the perspective of question, there are two settings, \textit{ answer source known} and \textit{answer source unknown}. With \textit{ answer source known}, we know which data source is queried given a question thus could use the corresponding prompts in the previous experiments.
% With \textit{answer source unknown}, we don't know which data source is queried, thus we use a unified prompt with the demonstration's format only includes the question. 
% We conduct experiments using a dynamic demo retrieval method without knowing the dataset's structure so the demonstration's format only includes the question, with or without merging all structure data.

% The experimental results are shown in Table~\ref{tab:mixedQA}. 
% \textit{w/o structure} is without knowing the structure of data.\textit{ w/o merge} utilize only own data while \textit{w/ merge} utilize all 5 datasets data. Dynamic acc is the acc on itself dataset in the dynamic retrieve. In \textit{w/o structure and w/o merge}, the effect is positively correlated with the dynamic demo acc. While in \textit{w/o structure and w/ merge}, the decline mainly relies on the mistaken alignment in merged data.

\begin{figure}
    \centering
    \includegraphics[width=0.47\textwidth]{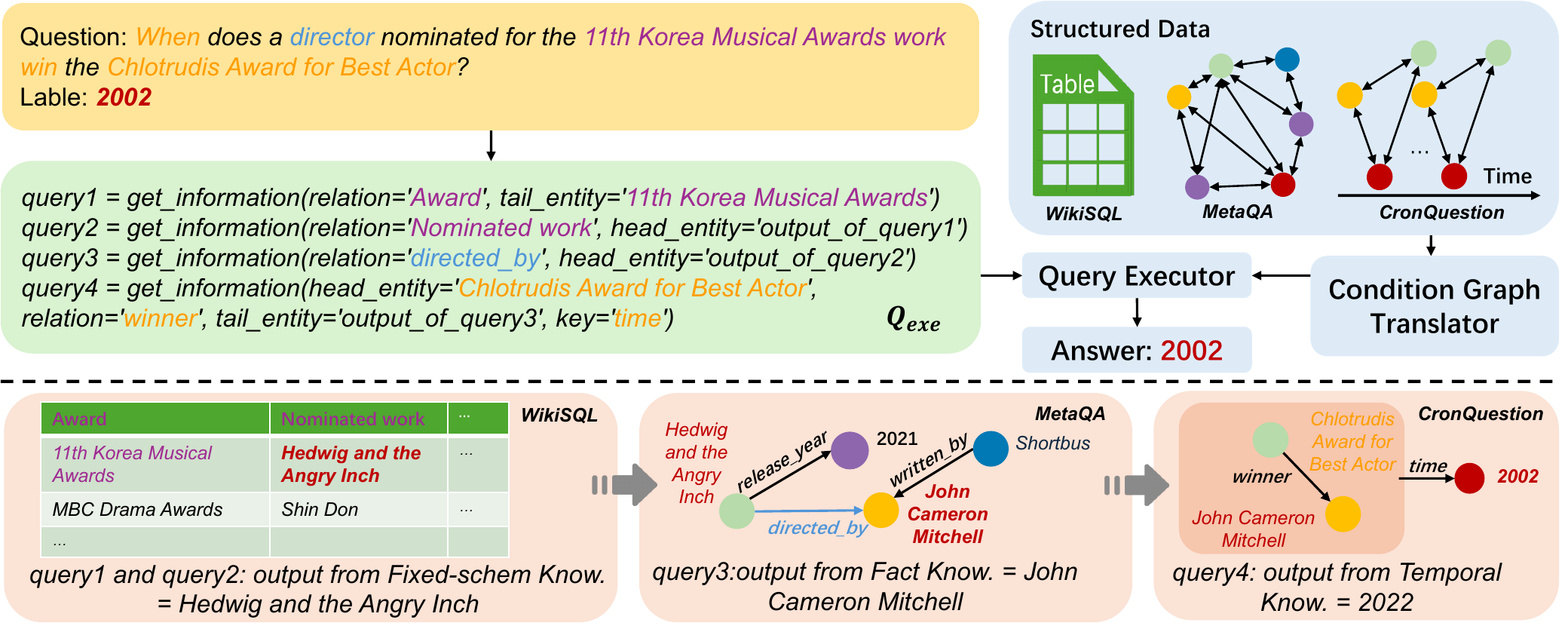}
    \vspace{-4mm}
    \caption{Case study of {\model} for across structured data. 
    % Above is the {\model} framework and below is the logical path in cross structure data.
    }
    \vspace{-3mm}
    \label{fig:case}
\end{figure}

\paragraph{QA across Structured Data}
    In this task, questions can be answered by relying on more than one data source. We construct a real case from 3 types of structure datasets: WikiSQL, MetaQA and CronQuestion. The case is shown in Figure~\ref{fig:case}. For question \textit{When does a director nominated for the 11st Korea Musical Awards work win the Chlotrudis Award for Best Actor?} $Q_{exe}$ first gets \textit{Hedwig and the Angry Inch} from a table in WikiSQL, then gets \textit{John Cameron Mitchell} from the KG in MetaQA, and finally gets the correct answer \textit{2002} from the temporal KG in CronQuestion. This case shows the potential of {\model} for QA across different types of structured data. We refer the reader to Appendix~\ref{exp:across} for more details of this experiment.

% analyze on cross knowledge
\vspace{-2mm}
\section{Discussion and Conclusion}
\label{sec:conclusion}
In this paper, we introduce a trustful framework for unified structured data QA, called {\model}, based on a novel and general data representation method named Condition Graph. Our method is interpretable as the LLM query and execution query are human-readable.  We experimentally prove the effectiveness of our method on 3 types of structured data and show the potential of dealing with more general and challenging scenarios. In the future, we will explore QA over mixed structured data and QA across structured data, both of which are more challenging but closer to real applications. 
% \input{secs/conclusion}

% \bibliography{aaai25}

\section{Acknowledgments}
This work is founded by National Natural Science Foundation of China (NSFC62306276/NSFCU23B2055/NSFCU19B2027), Zhejiang Provincial Natural Science Foundation of China (No. LQ23F020017), Yongjiang Talent Introduction Programme (2022A-238-G), and Fundamental Research Funds for the Central Universities (226-2023-00138). This work was supported by AntGroup.

\bibliography{aaai2025}

% \bibliography{aaai25}

\clearpage
\appendix 
\newpage

\section{Examples for Reasoning Function.}
\label{sec:examples4reasoning}
\begin{figure*}
    \centering
    % \vspace{-3mm}
\includegraphics[width=\textwidth]{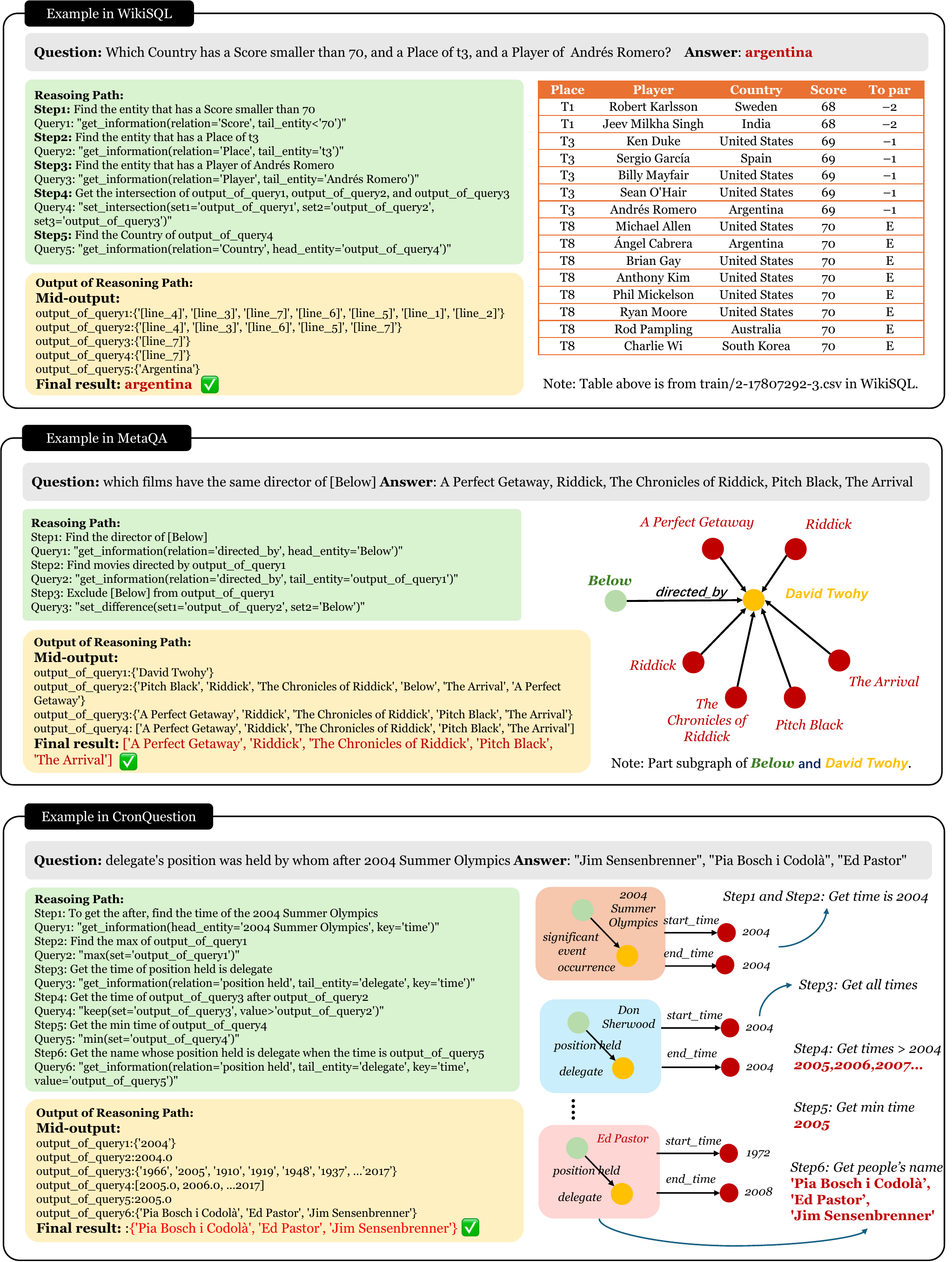}
% \vspace{-4mm}
    \caption{Examples for WikiSQL, MetaQA and CronQuestion.}
    \label{fig:examples}
    % \vspace{-3mm}
\end{figure*}

In Figure \ref{fig:examples}, we list 3 examples for using LLM Query $\mathcal{Q}_{llm}$ and set operations.

% \begin{figure*}
%     \centering
%     % \vspace{-3mm}
% \includegraphics[width=\textwidth]{figures/example_wikisql.pdf}
% % \vspace{-4mm}
%     \caption{Example in WikiSQL.}
%     \label{fig:{\model}}
%     % \vspace{-3mm}
% \end{figure*}
% % \vspace{-3mm}
% \begin{figure*}
%     \centering
% \includegraphics[width=\textwidth]{figures/example_metaqa.pdf}
% % \vspace{-4mm}
%     \caption{Example in MetaQA.}
%     \label{fig:{\model}}
% \end{figure*}
% % \vspace{-3mm}
% \begin{figure*}
%     \centering
% \includegraphics[width=\textwidth]{figures/example_cron.pdf}
% % \vspace{-4mm}
%     \caption{Example in MetaQA.}
%     \label{fig:{\model}}
% \end{figure*}
% \vspace{-3mm}

\section{Dataset Statistics}
\label{sec:datasets}
We conduct experiments on 5 datasets about 3 types of structured data.

We choose 2 table QA benchmarks, WikiSQL \cite{DBLP:journals/corr/abs-1709-00103} and WTQ \cite{wtq}, where WikiSQL is a large table QA dataset containing over 15 thousand test QA pairs, and WTQ demands more complex reasoning specifically related to tables.

We choose 2 KGQA benchmarks, WebQuestionsSP(WebQSP) \cite{DBLP:conf/acl/YihRMCS16} and MetaQA \cite{DBLP:conf/aaai/ZhangDKSS18}, which are constructed from Freebase and Wikidata. MetaQA includes 1-3 hops questions, thus questions included in MetaQA could be classified into  1 hop, 2 hops, and 3 hops. 

For temporal knowledge QA, we choose a recently widely used dataset CronQuestions  \cite{DBLP:conf/acl/SaxenaCT20}, which is constructed from temporal knowledge graph in Wikidata. According to the question type, the questions are classified into complex and simple questions. According to the answer type, the questions are classified into entity questions and time questions. 

The statistics are shown in Table~\ref{tab:datasets}.

\begin{table}[!hbpt]
% \begin{wraptable}{l}{0.7\textwidth}
    \centering
    \vspace{-3mm}
    \caption{Statistics of experimental datasets.}
    \resizebox{.45\textwidth}{!}{
    \begin{tabular}{l|c l | l c l}
    \toprule
     \textbf{Datasets}  & \textbf{\#test QA} &  \textbf{Others}  & \textbf{Datasets}  & \textbf{\#test QA} &  \textbf{Other Information} \\
    \midrule
      \textbf{WTQ}  &  4,344 & 421 tables & \textbf{WebQSP} & 1,639 & retrieved version\\
      \textbf{WikiSQL} & 15,878 & 5,230 tables & \textbf{CronQuestion}  & 30,000 & 5,000 triples\\
      \textbf{MetaQA} &39,093 & 134,741 triples & \quad - Simple & 17,142 & simple time and entity \\ 
      \quad - 1-hop & 9,947& simple 1 hop & \quad - Complex & 12,858 & three complex type\\
      \quad - 2-hop &14,872 & 21 types & \quad - Entity & 19,524 & query for entity\\
     \quad  - 3-hop &14,274 & 15 types & \quad - Time & 10,476& query for time\\
      % CronQuestion  & 30,000 & 5,000 triples\\
      % \quad - Complex & 12,858 & three complex type\\
      % \quad - Simple & 17,142 & simple time and entity\\
      % \quad - Entity & 19,524 & query for entity\\
      % \quad - Time & 10,476& query for time\\
      \bottomrule
    \end{tabular}
    }
    \label{tab:datasets}
% \end{wraptable}
\end{table}
% \section{Stronger LLMs on fixed-schema knowledge QA} \label{sec:gpt4}
% \input{tables/tableQA_gpt4}
% The results with GPT-4 are shown in Table \ref{tab:tableQA_gpt4}.
% In WikiSQL, {\model} outperformed UnifiedSKG and improved by 1.5 over GPT3.5 which comes to 87.4. 
% While in WTQ, the performance further improved, and outperformed UnifiedSKG with an accuracy of 52.1\%.
% This shows {\model}'s proficiency in deep semantic understanding, and indicates that {\model} will perform better as the ability of LLM itself improves.

\section{WebQSP Experiment Detail}
\label{sec:wqsp_detail}
\begin{table}[!hbpt]
\centering
\caption{The numbers of the semantic-clear and semantic-unclear in WebQSP.}
\resizebox{0.45\textwidth}{!}{
\label{tab:wqsp_semantic}
\begin{tabular}{c|ccc}
\hline
       & \textit{\# semantic-clear} & \textit{\# semantic-unclear} & \# all \\ \hline
WebQSP & 344              & 1295               & 1639  \\ \hline
\end{tabular}
}
\end{table}
For WebQSP, a relation is regarded as semantic-clear if there are a head entity and a tail entity in some of its associated triples that have meaning in natural language, such as ``ns:m.0xhsh" has meaning of ``Diamond".Otherwise, it is deemed a semantic-unclear relation. The numbers of the two types are shown in the Table \ref{tab:wqsp_semantic}.
We conduct separate experiments for each type and provide the overall results.

\begin{table}[!hbpt]
\centering
\caption{The experiment on WebQSP with semantic clear and unclear.}
\label{tab:wqsp_detail}
\resizebox{0.45\textwidth}{!}{
\begin{tabular}{c|ccc}
\hline Method & \textit{semantic-clear} & \textit{semantic-unclear} & all  \\ \hline
\textbf{\model}(ours) & 91.0\%          & 81.5\%            & 83.5\% \\ \hline
\end{tabular}
}
\end{table}

Experiment can be seen in Table \ref{tab:wqsp_detail}. Notably, we observe a significant difference between the results of questions with semantic-clear and -unclear relations, which are $91.0\%$  and $81.5\%$ respectively. Thus proper relation and entity naming in the question is important for understanding its meaning by the LLM.

\section{Potentials of TrustUQA Experiment Detail}
Since there is no existing mixed/across data dataset, we first constructed one using available datasets and then described the experiment in detail.

\subsection{QA over Mixed Structured Data}\label{exp:mixed}
\textbf{Dataset Construction:} We randomly selected 100 questions in each of WikiSQL, WTQ, MetaQA (1-hop), WebQSP, and CronQuestion datasets, extracting the corresponding structured data for each question. For the MetaQA (1-hop) and CronQuestion datasets, all the data is put together, we extracted only the one-hop data of the topic entity.
``Unmixed data" refers to the initial experience, where the structured data corresponding to each question is clearly defined.
``Mixed data" refers to the combination of structured data from the five constructed datasets into a single dataset, represented uniformly through a condition graph.

\textbf{Experimental Details:}
For ``answer source unknown", we cannot restrict the search to questions from the corresponding training set alone during dynamic retrieval; instead, we retrieve from all five training sets combined. Additionally, due to the lack of structured source, the demonstration format in the prompt includes only the question, without any auxiliary information like the topic entity and so on.

\subsection{QA across Structured Data Experiment Detail}\label{exp:across}
\begin{table*}[t!]
    \centering
    \resizebox{\textwidth}{!}{
    \begin{tabular}{ c l l l c }
        \hline
        \textbf{Year} & \textbf{Award} & \textbf{Category} & \textbf{Nominated work} & \textbf{Result} \\ \hline
        2000 & The National Theater Association of Korea & Best New Actor & Yi & Won \\ 
        2005 & 11th Korea Musical Awards & Most Popular Star & Hedwig and the Angry Inch & Won \\ 
        2005 & 11th Korea Musical Awards & Best Actor & Hedwig and the Angry Inch & Won \\
        2005 & MBC Drama Awards & Best New Actor & Shin Don & Nominated \\ 
        2006 & 12th Korea Musical Awards & Most Popular Star & Finding Kim Jong-wook & Won \\
        2006 & KBS Drama Awards & Best Couple Award with Yoon Eun-hye & The Vineyard Man & Won \\
        2006 & KBS Drama Awards & Popularity Award & The Vineyard Man & Won \\ 
        2006 & KBS Drama Awards & Best New Actor & The Vineyard Man & Won \\ 
        2007 & 1st The Musical Awards & Male Popularity Award & A Day & Won \\ 
        2007 & 13th Korea Musical Awards & Most Popular Star & A Day & Won \\ 
        2007 & SBS Drama Awards & Excellence Award, Actor in a Serial Drama & The King and I & Won \\ 
        2009 & KBS Drama Awards & Excellence Award, Actor in a Daily Drama & Everybody Cha-cha-cha & Won \\ \hline
    \end{tabular}
    }
    \vspace{0.7mm}
    \caption{test/2-11308224-9.csv in WikiSQL.}
    \label{tab:cross_table}
\end{table*}
\textbf{Dataset Construction:} 
Across Structured Data requires the construction of both structured data and corresponding questions. Structured data includes tables, knowledge graphs (KG), and temporal knowledge graphs (TKG). The construction approach follows "A->B->C->D," where the query A->B is the logic on table, B->C is the logic on KG, and C->D is the logic on TKG. This query logic indicates that B data appears in both the table and KG, C appears in both KG and TKG, and B and C appear together in a KG triple. Using these constraints, we can filter the data.

For our datasets, we chose WikiSQL for the table data, MetaQA for the KG, and CronQuestion for the TKG. Specifically, we used:
\begin{itemize}[leftmargin=0.2cm, itemindent=0.2cm]
    \item {WikiSQL}: test/2-11308224-9.csv can be seen at Table ~\ref{tab:cross_table}.
    \item {MetaQA}: triple at line 54952 (Hedwig and the Angry Inch directed\_by John Cameron Mitchell)
    \item {CronQuestion}: data at line 116380 (Chlotrudis Award for Best Actor winner John Cameron Mitchell \textbf{2002 2002})
\end{itemize}
Based on the above datasets, we constructed the question: ``When does a director nominated for the 11th Korea Musical Awards win the Chlotrudis Award for Best Actor?" The answer is \textbf{2002}.

\textbf{Experimental Details:}
We manually constructed the following execution query:
\begin{tcolorbox}
[title=Execution Query for QA across Structured Data] % colback
% \tiny
query1 = get\_information(relation=`Award', tail\_entity=`11th Korea Musical Awards')

query2 = get\_information(relation=`Nominated work', head\_entity=`output\_of\_query1')

query3 = get\_information(relation=`directed\_by', head\_entity=`output\_of\_query2')

query4 = get\_information(head\_entity=`Chlotrudis Award for Best Actor', relation=`winner', tail\_entity=`output\_of\_query3', key=`time')
\end{tcolorbox}
The result of execution query
 is \textbf{2002} matches the label, demonstrating the potential of {\model} in QA across Structured Data.

\section{Initial Demonstrations and LLM Prompts for Each Dataset}\label{sec:prompt}
The following presents static demonstrations and prompts on each dataset. The prompts follow the Chat Completions format, where ``role" includes ``system," ``user," and ``assistant". In the upcoming prompt displays, we prefix each role with \#\#\# to distinguish it from the content. Detailed results are as follows.
\subsection{WikiSQL}\label{sec:wikisql_prompt}
\paragraph{Initial Demonstrations}
WikiSQL has 8 demonstrations, the following is the detail.

\begin{lstlisting}[basicstyle = \scriptsize\ttfamily, breaklines = true]
### USER:
Schema: Pole Position:Robbie Kerr|Circuit:Oulton Park|Round:5|Fastest Lap:Heikki Kovalainen|Winning team:Carlin Motorsport|Winning driver:Robbie Kerr|Date:15 September.
Question: How many pole positions for round 20?
### ASSISTANT:
## Target type: {number}
## Query:
Step1: Find the pole positions for round 20
Query1: "get_information(relation='Pole Position', tail_entity='round 20')"
Step2: Calculate the count of output_of_query1
Query2: "count(set='output_of_query1')"

### USER:
Schema: Competition:2006 FIFA World Cup qualification|Date:October 8, 2005|Result:1-2|Score:4-2|Venue:Orange Bowl, Miami, United States|Goal:5.
Question: Which Competition has a Venue of estadio alfonso lastras, san luis potosi,mexico, and a Goal larger than 15?
### ASSISTANT:
## Target type: {competition}
## Query:
Step1: Find the entity that has venue of "estadio alfonso lastras, san luis potosi, mexico"
Query1: "get_information(relation='Venue', tail_entity='estadio alfonso lastras, san luis potosi, mexico')"
Step2: Find the entity that has goal larger than 15
Query2: "get_information(relation='Goal', tail_entity>'15')"
Step3: Get the intersection of output_of_query1 and output_of_query2
Query3: "set_intersection(set1 = 'output_of_query1', set2 = 'output_of_query2')"
Step4: Find the competition of output_of_query3
Query4: "get_information(relation='Competition', head_entity='output_of_query3')"

### USER:
Schema: College:Harding|Position:Tackle|NFL Team:Buffalo Bills|Player:Brett Miller|Pick #:135.
Question: What is the highest pick number the los angeles raiders got?
### ASSISTANT:
## Target type: {pick number}
## Query:
Step1: Find the entity that is the Los Angeles Raiders
Query1: "get_information(relation='NFL Team', tail_entity='Los Angeles Raiders')"
Step2: Find the highest pick number of output_of_query1
Query2: "get_information(relation='Pick #', head_entity='output_of_query1')"
Step3: Calculate max of output_of_query2
Query3: "max(set1='output_of_query2')"

### USER:
Schema: Name origin:Susan B.; American suffrage leader(1820-1906).|Diameter (km):169.0|Longitude:195.5E|Year named:1991|Name:Garland Patera|Latitude:6.5S.
Question: What is the longitude of the feature named Razia Patera?
### ASSISTANT:
## Target type: {longitude}
## Query:
Step1: Find the entity named Razia Patera
Query1: "get_information(relation='Name', tail_entity='Razia Patera')"
Step2: Find the longitude of output_of_query1
Query2: "get_information(relation='Longitude', head_entity='output_of_query1')"

### USER:
Schema: Winnings:$98,245|Driver:Scott Riggs|Make:Chevrolet|Car #:18|Laps:491|Points:155.
Question: What is the average car number of all the drivers who have won $111,683?
### ASSISTANT:
## Target type: {car number}
## Query:
Step1: Find the entity whose won is $111,683
Query1: "get_information(relation='Winnings', tail_entity='111,683')"
Step2: Find the car number of output_of_query1
Query2: "get_information(relation='Car #', head_entity='output_of_query1')"
Step3: Calculate the average of output_of_query2
Query3: "mean(set='output_of_query2')"

### USER:
Schema: Draws:3|Club:Real Jaen CF|Losses:20|Goal Difference:1|Goals for:37|Goals against:30|Wins:8|Points:15|Position:9|Played:30.
Question: How many Wins have Goals against smaller than 30, and Goals for larger than 25, and Draws larger than 5?
### ASSISTANT:
## Target type: {number of wins}
## Query:
Step1: Find the Wins with Goals against smaller than 30
Query1: "get_information(relation='Goals against', tail_entity<'30')"
Step2: Find the Wins with Goals for larger than 25
Query2: "get_information(relation='Goals for', tail_entity>'25')"
Step3: Find the Wins with Draws larger than 5
Query3: "get_information(relation='Draws', tail_entity>'5')"
Step4: Get the intersection of output_of_query1, output_of_query2, and output_of_query3
Query4: "set_intersection(set1='output_of_query1', set2='output_of_query2', set3='output_of_query3')"
Step5: Get the count of output_of_query4
Query5: "count(set='output_of_query4')"

### USER:
Schema: Record Set:9 nominations|Year:1967|Actor:Henry Fonda|Notes:Skippy|Superlative:Oldest Winner.
Question: What year was the the youngest nominee a winner?
### ASSISTANT:
## Target type: {year}
## Query:
Step1: Find the youngest nominee who was a winner
Query1: "get_information(relation='Superlative', tail_entity='youngest nominee')"
Step2: Get the year of output_of_query1
Query2: "get_information(relation='Year', head_entity='output_of_query1')"

### USER:
Schema: Season:1965/66|Name:Luis Arconada|minutes:709|Club:Atletico Madrid|Rank:2.
Question: What are the minutes of the Player from Real Madrid Club with a Rank of 7 or larger?
### ASSISTANT:
## Target type: {minutes}
## Query:
Step1: Find the entity that belongs to Real Madrid Club
Query1: "get_information(relation='Club', tail_entity='Real Madrid Club')"
Step2: Find the entity with a Rank of 7 or larger
Query2: "get_information(relation='Rank', tail_entity>='7')"
Step3: Get the intersection of output_of_query1 and output_of_query2
Query3: "set_intersection(set1 = 'output_of_query1', set2 = 'output_of_query2')"
Step4: Get the minutes of output_of_query3
Query4: "get_information(relation='minutes', head_entity='output_of_query3')"
\end{lstlisting}

\paragraph{LLM Prompt} LLM Prompt is the description for conditional graph query, following is the detail.

\begin{lstlisting}[basicstyle = \scriptsize\ttfamily, breaklines = true]
### SYSTEM:
Now you are learning to write conditional graph queries for natural language questions. 
And conditional graph query function is:
def get_information(relation, tail_entity, head_entity, key, value):
"""
    This function retrieves information by querying a data source using a given relation and tail entity as search criteria.
    Args:
        relation (str): The relation to the query that matches the tail_entity or contains the  head_entity.
        tail_entity (str): The tail entity associated with the relation.
        head_entity (str): The head entity belongs to the relation.
        key (str):  The key to query that matches the tail_entity or head_entity.
        value (str):  The value associated with or belongs to the key.
    Returns:
        list: A list of query results.
    """
To finish the question correctly, you can combine the Set and Calculator functions.

Set functions:
- set_union(set1, set2)
- set_intersection(set1, set2)
- set_difference(set1, set2)
- set_negation(set)
...
Your task is to break down the original natural language problem step by step with given all relation types and a corresponding example as schema with the format of "relation:example". Following are some examples.
Note: in get_information function, tail_entity and head_entity cannot occur together.
\end{lstlisting}

\subsection{WTQ}\label{sec:wtq_prompt}
\paragraph{Initial Demonstrations}
WTQ has 8 demonstrations, the following is the detail.
\begin{lstlisting}[basicstyle = \scriptsize\ttfamily, breaklines = true]
### USER:
Schema: Date:14 February 1976|Tie no:Replay|row_number:6|Score:1-2|Home team:Derby County|Away team:Crystal Palace.
Question: who was the home team in the game on the top of the table?
### ASSISTANT:
## Target type: {home team}
## Query:
Step1: Find the game on the top of the table
Query1: "get_information(relation='row_number', tail_entity='1')"
Step2: Get the home team of output_of_query1
Query2: "get_information(relation='Home team', tail_entity='output_of_query1')"

### USER:
Schema: Fastest Lap:Troy Corser|Report:Report|Pole Position:Troy Bayliss|Round:R2|Circuit:Losail|row_number:5|Date:3 August|Country:Europe|Winning Team:Yamaha Motor Italia|Winning Rider:Noriyuki Haga.
Question: who did the most pole position what is his or her name
### ASSISTANT:
## Target type: {'Pole Position'}
## Query:
Step1: Find the entity in Pole Position that appears the most.
Query1: "get_information(relation='Pole Position', tail_entity>='most')"
Step2: Find the Pole Position of output_of_query1.
Query2: "get_information(relation='Pole Position', head_entity='output_of_query1')"

### USER:
Schema: Silver:0|Gold:0|Bronze:2|Nation:Venezuela\xa0(VEN)|row_number:5|Rank:7|Total:1.
Question: how many nations scored more total medals than brazil?
### ASSISTANT:
## Target type: {number of nations}
## Query:
Step1: Find the entity that is Brazil
Query1: "get_information(relation='Nation', tail_entity='Brazil')"
Step2: Find the total medals of output_of_query1
Query2: "get_information(relation='Total', head_entity='output_of_query1')"
Step3: Find the entity whose total medals is larger than output_of_query2
Query3: "get_information(relation='Total', tail_entity>'output_of_query2')"
Step4: Calculate the count of output_of_query3
Query4: "count(set='output_of_query3')"

### USER:
Schema: Host:University of Texas|Rnd:1|Venue:Thomas Assembly Center|State:Maryland|City:Norfolk|row_number:16|Region:East.
Question: total number of venues that hosted first and second round games
### ASSISTANT:
## Target type: {number}
## Query:
Step1: Find the venues that hosted first round games
Query1: "get_information(relation='Rnd', tail_entity='1')"
Step2: Find the venues that hosted second round games
Query2: "get_information(relation='Rnd', tail_entity='2')"
Step3: Get the union of output_of_query1 and output_of_query2
Query3: "set_union(set1='output_of_query1', set2='output_of_query2')"
Step4: Calculate the count of output_of_query3
Query4: "count(set='output_of_query3')"

### USER:
Schema: row_number:9|No. in\\nseries:142|Written by:Adam Markowitz & Tom Devanney|Original air date:April\xa015,\xa01996|Title:"I, Bowl Buster"|Production\\ncode:60108|Directed by:Shelley Jensen|No. in\\nseason:12.
Question: what was the last episode called?
### ASSISTANT:
## Target type: {'Title'}
## Query:
Step1: Find all the row_number
Query1: "get_information(relation='row_number')"
Step2: Find the max row_number
Query2: "max(set='output_of_query1')"
Step3: Get the row with max row_number 
Query3: "get_information(relation='row_number', tail_entity='output_of_query2')"
Step4: Get the Title of output_of_query3
Query4: "get_information(relation='Title', head_entity='output_of_query3')"

### USER:
Schema: Number of Votes:60,048|Winner:James A. Leach|row_number:46|Party Affiliation:Republican|Loser:John R. Schmidhauser|Year:1946|Percentage of Votes:50% - 49%.
Question: what winner received the least number of votes?
### ASSISTANT:
## Target type: {country}
## Query:
Step1: Find the entity with the relation of Number of Votes
Query1: "get_information(relation='Number of Votes')"
Step2: Get the min output_of_query1
Query2: "min(set='output_of_query1')" 
Step3: Get the entity with the min Votes
Query3: "get_information(relation='Number of Votes', tail_entity='output_of_query2')"
Step4: Find the winner of output_of_query3
Query4: "get_information(relation='Winner', head_entity='output_of_query3')"

### USER:
Schema: Islam:96.67%|n/a:1.16%|Other:0.04%|None\xa0/ Atheism:6.09%|Christianity:0.39%|Buddhism:0.01%|Ethnicity:Kyrgyz|Judaism:0.03%|row_number:13.
Question: which ethnicity is previous from dungan'
### ASSISTANT:
## Target type: {ethnicity}
## Query:
Step1: Find the entity that Ethnicity is Dungan
Query1: "get_information(relation='Ethnicity', tail_entity='Dungan')"
Step2: Find the row_number of output_of_query1
Query2: "get_information(relation='row_number', head_entity='output_of_query1')"
Step3: Find the previous_row of output_of_query2
Query3: "previous_row(set='output_of_query2')"
Step4: Find the row_number is output_of_query3
Query4: "get_information(relation='row_number', tail_entity='output_of_query3')"
Step5: Find the Ethnicity of output_of_query4
Query5: "get_information(relation='Ethnicity', head_entity='output_of_query4')"

### USER:
Schema: Residence:Morris Plains|row_number:68|Name:Jerry Green|Party:Rep|First served:2008|District:District 31.
Question: angel fuentes and gilbert wilson were both from which residence?
### ASSISTANT:
## Target type: {residence}
## Query:
Step1: Find the entity named Angel Fuentes
Query1: "get_information(relation='Name', tail_entity='Angel Fuentes')"
Step2: Find the residence of output_of_query1
Query2: "get_information(relation='Residence', head_entity='output_of_query1')"
Step3: Find the entity named Gilbert Wilson
Query3: "get_information(relation='Name', tail_entity='Gilbert Wilson')"
Step4: Find the residence of output_of_query3
Query4: "get_information(relation='Residence', head_entity='output_of_query3')"
Step5: Get the intersection of output_of_query2 and output_of_query4
Query5: "set_intersection(set1='output_of_query2', set2='output_of_query4')"

\end{lstlisting}

\paragraph{LLM Prompt}
LLM Prompt is the description for conditional graph query, following is the detail.

\begin{lstlisting}[basicstyle = \scriptsize\ttfamily, breaklines = true]
### SYSTEM:
Now you are learning to write conditional graph queries for natural language questions. 
And conditional graph query function is:
def get_information(relation, tail_entity, head_entity, key, value):
"""
    This function retrieves information by querying a data source using a given relation and tail entity as search criteria.
    
    Args:
        relation (str): The relation to the query that matches the tail_entity or contains the  head_entity.
        tail_entity (str): The tail entity associated with the relation.
        head_entity (str): The head entity belongs to the relation.
        key (str):  The key to query that matches the tail_entity or head_entity.
        value (str):  The value associated with or belongs to the key.
        
    Returns:
        list: A list of query results.
    """
To finish the question correctly, you can combine the Set and Calculator functions.
Set functions:
- set_union(set1, set2)
- set_intersection(set1, set2)
- set_difference(set1, set2)
- set_negation(set1)
...
You also can use previous_row and next_row two functions to find the previous or next information.
Your task is to break down the original natural language problem step by step with given all relation types and a corresponding example as schema with the format of "relation:example". Following are some examples.
Note: in get_information function, tail_entity and head_entity cannot occur together.
\end{lstlisting}

\subsection{WebQSP}\label{sec:wqsp_prompt}
\underline{\textbf{WebQSP with semantic clear relations}}: The prompt for WebQSP with semantic clear relations has 4 demonstrations and the description for conditional graph query, following is the detail.
\paragraph{Initial Demonstrations} WebQSP semantic-clear has 4 demonstrations, the following is the detail.

\begin{lstlisting}[basicstyle = \scriptsize\ttfamily, breaklines = true]
### USER:
Question: what organization did dr. carter g. woodson found?
TopicEntityName: ['Carter G. Woodson']
First_step: ['Carter G. Woodson', 'organization.organization_founder.organizations_founded', 'Association for the Study of African American Life and History']|['Carter G. Woodson', 'people.person.nationality', 'United States of America']
Second_step: ['Carter G. Woodson', 'organization.organization_founder.organizations_founded', 'Association for the Study of African American Life and History']|['Carter G. Woodson', 'people.person.nationality', 'United States of America']|['African American', 'book.book_subject.works', "Looking at the World through Michael Jackson's Left Eye (Part 2)"]|['Mobile', 'location.location.people_born_here', 'JaMarcus Russell']|['United Kingdom', 'location.country.second_level_divisions', 'Midlothian']|['Montgomery Improvement Association', 'organization.organization.founders', 'Martin Luther King, Jr.']|['Austin', 'base.biblioness.bibs_location.country', 'United States of America']|['United States of America', 'government.governmental_jurisdiction.governing_officials', 'Mark Takano']|['Jamaica Island', 'location.location.contains', 'Cockpit Country']
### ASSISTANT:
Step1: Get the organizations founded by Carter G. Woodson
Query1: "get_information(relation='organization.organization_founder.organizations_founded', head_entity='Carter G. Woodson')"

### USER:
Question: what did ronnie radke do?
TopicEntityName: ['Ronnie Radke']
First_step: ['Ronnie Radke', 'people.person.profession', 'Singer-songwriter']
Second_step: ['Ronnie Radke', 'people.person.profession', 'Singer-songwriter']|['Mobile', 'location.location.people_born_here', 'JaMarcus Russell']|['Dance music', 'music.genre.albums', 'Number Ones']|['United States of America', 'government.governmental_jurisdiction.governing_officials', 'Mark Takano']|['Jamaica Island', 'location.location.contains', 'Cockpit Country']
### ASSISTANT:
Step1: Get the profession of Ronnie Radke
Query1: "get_information(relation='people.person.profession', head_entity='Ronnie Radke')"

### USER:
Question: where is downtown littleton?
TopicEntityName: ['Littleton, Colorado']
First_step: ['Arapahoe County', 'location.location.time_zones', 'Mountain Time Zone']|['Littleton', 'location.location.containedby', 'United States of America']
Second_step: ['Arapahoe County', 'location.location.time_zones', 'Mountain Time Zone']|['Littleton', 'location.location.containedby', 'United States of America']|['Central Time Zone', 'time.time_zone.locations_in_this_time_zone', 'Merritt, Illinois']|['Denmark', 'topic_server.population_number', '"3835"']|['Bob Marley', 'people.person.gender', 'Male']|['Australian dollar', 'common.topic.alias', '"AUD"']|['United States of America', 'government.governmental_jurisdiction.governing_officials', 'Mark Takano']|['Jamaica Island', 'location.location.contains', 'Cockpit Country']
### ASSISTANT:
Step1: Get the containedby location of Littleton, Colorado
Query1: "get_information(relation='location.location.containedby', head_entity='Littleton, Colorado')"

### USER:
Question: what is south america made up of?
TopicEntityName: ['South America']
First_step: ['South America', 'location.location.contains', 'Loncomilla River']
Second_step: ['South America', 'location.location.contains', 'Loncomilla River']|['Saint Thomas Parish', 'location.location.partially_contains', 'Blue Mountain Peak']|['Turner Field', 'location.location.events', '1999 World Series']|['Mississippi River', 'location.location.partially_containedby', 'Illinois']|['Roman Empire', 'location.country.official_language', 'Latin Language']|['Jamaica', 'location.country.languages_spoken', 'Jamaican Creole English Language']|['France', 'sports.sports_team_location.teams', 'France national baseball team']|['Prison Oval', 'common.topic.notable_types', 'Structure']|['Egypt', 'location.country.form_of_government', 'Semi-presidential system']|['jm', 'type.object.name', '"jm"']|['Franklin', 'symbols.namesake.named_after', 'Benjamin Franklin']|['United States of America', 'government.governmental_jurisdiction.governing_officials', 'Mark Takano']|['Lucea West River', 'location.location.containedby', 'Jamaica']
### ASSISTANT:
Step1: Get the components that make up South America
Query1: "get_information(relation='location.location.contains', head_entity='South America')"
\end{lstlisting}
\paragraph{LLM Prompt} LLM Prompt is the description for conditional graph query, following is the detail.

\begin{lstlisting}[basicstyle = \scriptsize\ttfamily, breaklines = true]
### SYSTEM:
Now you are learning to write conditional graph queries for natural language questions. 
And conditional graph query function is:
def get_information(relation, tail_entity, head_entity):
"""
    This function retrieves information by querying a data source using a given relation and tail entity as search criteria.
    Args:
        relation (str): The relation to the query that matches the tail_entity or contains the  head_entity.
        tail_entity (str): The tail entity associated with the relation in TopicEntityName.
        head_entity (str): The head entity belongs to the relation in TopicEntityName.
    Returns:
        list: A list of query results.
    """
Your task is to use the relation in First_step in the first step and Second_step for second (if necessary) in which the format is given with [head, relation, tail].
Note: 
1. In get_information function, tail_entity and head_entity cannot occur together.
\end{lstlisting}

\underline{\textbf{WebQSP with semantic unclear relations}}: The prompt for WebQSP with semantic unclear relations has 6 demonstrations and the description for conditional graph query, the following is the detail.
\paragraph{LLM Prompt} WebQSP emantic-unclear has 6 demonstrations, the following is the detail.

\begin{lstlisting}[basicstyle = \scriptsize\ttfamily, breaklines = true]
### USER:
Question: where is the nation of palestine?
TopicEntityName: ['Palestine']
First_step: ['location.location.contains', 'location.location.containedby']
Second_step: ['base.locations.continents.countries_within', 'location.location.contains', 'location.location.containedby', 'location.location.partially_contains', 'base.biblioness.bibs_location.country', 'location.location.events']
### ASSISTANT:
Step1: Get the nation of Palestine
Query1: "get_information(relation='location.location.containedby', head_entity='Palestine')"

### USER:
Question: who does nikki reed play in the movie twilight?
TopicEntityName: ['Nikki Reed']
First_step: ['film.actor.film']
Second_step: ['government.politician.government_positions_held', 'topic_server.population_number', 'film.film.country', 'film.actor.film', 'people.person.place_of_birth', 'film.film.directed_by', 'sports.pro_athlete.career_start', 'people.person.profession', 'film.performance.character', 'music.artist.genre', 'people.deceased_person.place_of_death', 'film.film.production_companies', 'location.location.time_zones', 'people.person.religion', 'film.film.starring', 'people.person.nationality', 'travel.tourist_attraction.near_travel_destination', 'common.topic.notable_types', 'government.government_position_held.governmental_body', 'film.film.language', 'film.performance.film', 'film.film.genre', 'book.written_work.subjects', 'location.location.containedby']
### ASSISTANT:
Step1: Get the film that Nikki Reed acted in
Query1: "get_information(relation='film.actor.film', head_entity='Nikki Reed')"
Step2: Get the character name of output_of_query1
Query2: "get_information(relation='film.performance.character', head_entity=output_of_query1)"

### USER:
Question: where did chris cuomo go?
TopicEntityName: ['Andrew Cuomo']
First_step: ['people.person.education']
Second_step: ['government.politician.government_positions_held', 'topic_server.population_number', 'film.film.country', 'location.location.containedby', 'people.person.profession', 'people.person.education', 'sports.pro_athlete.teams', 'people.deceased_person.place_of_death', 'people.person.nationality', 'travel.tourist_attraction.near_travel_destination', 'music.artist.track_contributions', 'common.topic.notable_types', 'government.government_position_held.governmental_body', 'location.location.contains', 'government.government_office_or_title.office_holders', 'education.education.institution', 'film.film.starring', 'government.government_position_held.office_holder']
### ASSISTANT:
Step1: Get the education of Andrew Cuomo
Query1: "get_information(relation='people.person.education', head_entity='Andrew Cuomo')"
Step2: Get the institution of output_of_query1
Query2: "get_information(relation='education.education.institution', head_entity=output_of_query1)"

### USER:
Question: who was the leader of the ussr in 1948?
TopicEntityName: ['Soviet Union']
First_step: ['location.location.events', 'organization.organization_founder.organizations_founded', 'common.topic.notable_types', 'government.governmental_jurisdiction.governing_officials', 'location.location.contains']
Second_step: ['government.politician.government_positions_held', 'location.location.adjoin_s', 'topic_server.population_number', 'base.locations.countries.continent', 'location.country.form_of_government', 'base.biblioness.bibs_location.loc_type', 'organization.organization_founder.organizations_founded', 'book.author.works_written', 'location.country.iso_numeric', 'location.location.events', 'common.topic.notable_types', 'government.government_office_category.officeholders', 'location.location.contains', 'government.government_position_held.basic_title', 'base.activism.activist.area_of_activism', 'government.government_position_held.jurisdiction_of_office', 'government.governmental_jurisdiction.governing_officials', 'location.country.capital', 'government.government_position_held.office_holder']
### ASSISTANT:
Step1: Get the governing_officials of Soviet Union
Query1: "get_information(relation='government.governmental_jurisdiction.governing_officials', head_entity='Soviet Union')"
Step2: Get the office_holder of output_of_query1
Query2: "get_information(relation='government.government_position_held.office_holder', head_entity=output_of_query1)"

### USER:
Question: who does jr smith play for?
TopicEntityName: ['J. R. Smith']
First_step: ['sports.pro_athlete.teams']
Second_step: ['film.film.music', 'topic_server.population_number', 'film.film.country', 'people.person.place_of_birth', 'location.location.containedby', 'music.album.release_date', 'sports.sports_league.teams', 'people.person.profession', 'sports.pro_athlete.teams', 'music.artist.genre', 'government.government_position_held.to', 'sports.sports_team_roster.team', 'location.location.time_zones', 'people.person.religion', 'people.person.nationality', 'film.writer.film', 'media_common.netflix_title.netflix_genres', 'common.topic.notable_types', 'film.film.language', 'government.government_position_held.jurisdiction_of_office', 'fictional_universe.fictional_character.character_created_by', 'base.culturalevent.event.entity_involved', 'government.political_district.representatives', 'film.film.starring']
### ASSISTANT:
Step1: Get the teams of J. R. Smith
Query1: "get_information(relation='sports.pro_athlete.teams', head_entity='J. R. Smith')"
Step2: Get roster of output_of_query1
Query2: "get_information(relation='sports.sports_team_roster.team', head_entity=output_of_query1)"

### USER:
Question: who did heinrich himmler marry?
First_step: ['people.person.spouse_s', 'base.activism.activist.area_of_activism']
Second_step: ['people.marriage.type_of_union', 'government.politician.government_positions_held', 'people.person.place_of_birth', 'people.person.spouse_s', 'sports.pro_athlete.career_start', 'music.group_member.instruments_played', 'people.person.profession', 'symbols.namesake.named_after', 'people.person.languages', 'people.deceased_person.place_of_death', 'people.person.children', 'people.person.religion', 'influence.influence_node.influenced', 'influence.influence_node.influenced_by', 'people.person.nationality', 'people.marriage.spouse', 'people.person.date_of_birth', 'sports.sports_team_owner.teams_owned']
### ASSISTANT:
Step1: Get the Person_Spouse of Heinrich Himmler
Query1: "get_information(relation='people.person.spouse_s', head_entity='Heinrich Himmler')"
Step2: Get name of output_of_query1
Query2: "get_information(relation='people.marriage.spouse', head_entity=output_of_query1)"
\end{lstlisting}
\paragraph{LLM Prompt} LLM Prompt is the description for conditional graph query, following is the detail.

\begin{lstlisting}[basicstyle = \scriptsize\ttfamily, breaklines = true]
### SYSTEM:
Now you are learning to write conditional graph queries for natural language questions. 
And conditional graph query function is:
def get_information(relation, tail_entity, head_entity):
"""
    This function retrieves information by querying a data source using a given relation and tail entity as search criteria.
    Args:
        relation (str): The relation to the query that matches the tail_entity or contains the  head_entity.
        tail_entity (str): The tail entity.
        head_entity (str): The head entity.
    Returns:
        list: A list of query results.
    """
Your task is to use the relation in First_step in the first step and Second_step for second (if necessary).
Note: 
1. In get_information function, tail_entity and head_entity cannot occur together.
\end{lstlisting}

\subsection{MetaQA}\label{sec:meta_prompt}
\underline{\textbf{MetaQA 1-hop}}: The prompt for MetaQA 1-hop has 13 demonstrations and the description for conditional graph query, following is the detail.
\paragraph{Initial Demonstrations}
MetaQA 1-hop has 13 demonstrations, the following is the detail.

\begin{lstlisting}[basicstyle = \scriptsize\ttfamily, breaklines = true]
### USER:
Question: what movies did [Temuera Morrison] act in
### ASSISTANT:
Step1: Find the movies that [Temuera Morrison] acted in
Query1: "get_information(relation='starred_actors', tail_entity='Temuera Morrison')"

### USER:
Question: what films did [Charles Bail] direct
### ASSISTANT:
Step1: Find movies that [Charles Bail] directs
Query1: "get_information(relation='directed_by', tail_entity='Charles Bail')"

### USER:
Question: who acted in [Red River]
### ASSISTANT:
Step1: Find starred_actors in [Red River]
Query1: "get_information(relation='starred_actors', head_entity='Red River')"

### USER:
Question: the director of [Cube Zero] was who
### ASSISTANT:
Step1: Find director of [Cube Zero]
Query1: "get_information(relation='directed_by', head_entity='Cube Zero')"

### USER:
Question: what is the genre of the film [The Darjeeling Limited]
### ASSISTANT:
Step1: Find genre of [The Darjeeling Limited]
Query1: "get_information(relation='has_genre', head_entity='The Darjeeling Limited')"

### USER:
Question: how would people rate [Little Women]
### ASSISTANT:
Step1: Find rate of [Little Women]
Query1: "get_information(relation='has_imdb_rating', head_entity='Little Women')"

### USER:
Question: what is the primary language in [The Criminal Life of Archibaldo de la Cruz]
### ASSISTANT:
Step1: Find language of [The Criminal Life of Archibaldo de la Cruz]
Query1: "get_information(relation='in_language', head_entity='The Criminal Life of Archibaldo de la Cruz')"

### USER:
Question: what are some words that describe movie [The Omega Man]
### ASSISTANT:
Step1: Find tags of [The Omega Man]
Query1: "get_information(relation='has_tags', head_entity='The Omega Man')"

### USER:
Question: who's the writer of [Experiment Perilous]
### ASSISTANT:
Step1: Find tags of [The Omega Man]
Query1: "get_information(relation='written_by', head_entity='Experiment Perilous')"

### USER:
Question: when was the movie [BUtterfield 8] released
### ASSISTANT:
Step1: Find released year of [BUtterfield 8]
Query1: "get_information(relation='release_year', head_entity='BUtterfield 8')"

### USER:
Question: which films can be described by [romeo and juliet]
### ASSISTANT:
Step1: Find films which has tags of [romeo and juliet]
Query1: "get_information(relation='has_tags', tail_entity='romeo and juliet')"

### USER:
Question: which movies was [Jeff Stanzler] the writer of
### ASSISTANT:
Step1: Find films whose writer is [Jeff Stanzler]
Query1: "get_information(relation='written_by', tail_entity='Jeff Stanzler')"

### USER:
Question: what films are about [anthony's mackie]
### ASSISTANT:
Step1: Find films which has tags of [anthony's mackie]
Query1: "get_information(relation='has_tags', tail_entity='anthony's mackie')"

\end{lstlisting}
\paragraph{LLM Prompt} LLM Prompt is the description for conditional graph query, following is the detail.

\begin{lstlisting}[basicstyle = \scriptsize\ttfamily, breaklines = true]
### SYSTEM:
Now you are learning to write conditional graph queries for natural language questions. 
The conditional graph query function is:
def get_information(relation, tail_entity, head_entity, key, value):
"""
    This function retrieves information by querying a data source using a given relation and tail entity as search criteria.
    Args:
        relation (str): The relation to the query that matches the tail_entity or contains the  head_entity which includes {'directed_by', 'has_imdb_rating', 'release_year', 'in_language', 'written_by', 'has_tags', 'has_imdb_votes', 'has_genre', 'starred_actors'}.
        tail_entity (str): The tail entity associated with the relation.
        head_entity (str): The head entity belongs to the relation.
        key (str):  The key to query that matches the tail_entity or head_entity.
        value (str):  The value associated with or belongs to the key.
        
    Returns:
        list: A list of query results.
    """
To finish the question correctly, you can combine the Set and Calculator functions.
Set functions:
- set_union(set1, set2)
- set_intersection(set1, set2)
- set_difference(set1, set2)
- set_negation(set1)
...
Note: 
1. In get_information function, tail_entity and head_entity cannot occur together.
\end{lstlisting}
\underline{\textbf{MetaQA 2-hop}}: The prompt for MetaQA 2-hop has 15 demonstrations and the description for conditional graph query, following is the detail.
\paragraph{Initial Demonstrations}
MetaQA 2-hop has 15 demonstrations, the following is the detail.

\begin{lstlisting}[basicstyle = \scriptsize\ttfamily, breaklines = true]
### USER:
Question: which films share the same actor of [The Don't Amityville Curse]
### ASSISTANT:
Step1: Find the actors of [The Don't Amityville Curse]
Query1: "get_information(relation='starred_actors', head_entity='The Don't Amityville Curse')"
Step2: Find movies that output_of_query1 acted in
Query2: "get_information(relation='starred_actors', tail_entity='output_of_query1')"
Step3: Exclude [The Don't Amityville Curse] from output_of_query2
Query3: "set_difference(set1='output_of_query2', set2='The Don't Amityville Curse')"

### USER:
Question: the director of [The Slugger's Wife] is also the director of which movies
### ASSISTANT:
Step1: Find the director of [The Slugger's Wife]
Query1: "get_information(relation='directed_by', head_entity='The Slugger's Wife')"
Step2: Find movies that output_of_query1 directed
Query2: "get_information(relation='directed_by', tail_entity='output_of_query1')"
Step3: Exclude [The Slugger's Wife] from output_of_query2
Query3: "set_difference(set1='output_of_query2', set2='The Slugger's Wife')"

### USER:
Question: the screenwriter of [Taxidermia] also wrote which films
### ASSISTANT:
Step1: Find the writer of [Taxidermia]
Query1: "get_information(relation='written_by', head_entity='Taxidermia')"
Step2: Find movies that output_of_query1 wrote
Query2: "get_information(relation='written_by', tail_entity='output_of_query1')"
Step3: Exclude [Taxidermia] from output_of_query2
Query3: "set_difference(set1='output_of_query2', set2='Taxidermia')"

### USER:
Question: who acted together/co-star with [Tania Raymonde]
### ASSISTANT:
Step1: Find the movies that [Tania Raymonde] acted in
Query1: "get_information(relation='starred_actors', tail_entity='Tania Raymonde')"
Step2: Find actors of output_of_query1
Query2: "get_information(relation='starred_actors', head_entity='output_of_query1')"
Step3: Exclude [Tania Raymonde] from output_of_query2
Query3: "set_difference(set1='output_of_query2', set2='Tania Raymonde')"

### USER:
Question: who are the directors of the movies acted by [Khalil Kain]
### ASSISTANT:
Step1: Find the movies that [Khalil Kain] acted in
Query1: "get_information(relation='starred_actors', tail_entity='Khalil Kain')"
Step2: Find directors of output_of_query1 
Query2: "get_information(relation='directed_by', head_entity='output_of_query1')"
Step3: Exclude [Khalil Kain] from output_of_query2
Query3: "set_difference(set1='output_of_query2', set2='Khalil Kain')"

### USER:
Question: what genres do the movies acted by [Ulrich M\u00fche] fall under
### ASSISTANT:
Step1: Find the movies that [Ulrich M\u00fche] acted in
Query1: "get_information(relation='starred_actors', tail_entity='Ulrich M\u00fche')"
Step2: Find genres of output_of_query1
Query2: "get_information(relation='has_genre', head_entity='output_of_query1')"

### USER:
Question: the films acted by [Guglielmo Barnab\u00f2] were written by who
### ASSISTANT:
Step1: Find the movies that [Guglielmo Barnab\u00f2] acted in
Query1: "get_information(relation='starred_actors', tail_entity='Guglielmo Barnab\u00f2')"
Step2: Find the writer of output_of_query1
Query2: "get_information(relation='written_by', head_entity='output_of_query1')"
Step3: Exclude [Guglielmo Barnab\u00f2] from output_of_query2
Query3: "set_difference(set1='output_of_query2', set2='Guglielmo Barnab\u00f2')"

### USER:
Question: which directors co-directed films with [Graham Baker]
### ASSISTANT:
Step1: Find the movies that [Graham Baker] directed
Query1: "get_information(relation='directed_by', tail_entity='Graham Baker')"
Step2: Find the director of output_of_query1
Query2: "get_information(relation='directed_by', head_entity='output_of_query1')"
Step3: Exclude [Graham Baker] from output_of_query2
Query3: "set_difference(set1='output_of_query2', set2='Graham Baker')"

### USER:
Question: what genres do the movies directed by [Clark Gregg] fall under
### ASSISTANT:
Step1: Find the movies that [Clark Gregg] directed
Query1: "get_information(relation='directed_by', tail_entity='Clark Gregg')"
Step2: Find genres of output_of_query1
Query2: "get_information(relation='has_genre', head_entity='output_of_query1')"

### USER:
Question: what are the languages spoken in the films directed by [Joel Zwick]
### ASSISTANT:
Step1: Find the movies that [Joel Zwick] directed
Query1: "get_information(relation='directed_by', tail_entity='Joel Zwick')"
Step2: Find language of output_of_query1
Query2: "get_information(relation='in_language', head_entity='output_of_query1')"

### USER:
Question: when were the films directed by [David Green] released
### ASSISTANT:
Step1: Find the movies that [David Green] directed
Query1: "get_information(relation='directed_by', tail_entity='David Green')"
Step2: Find release year of output_of_query1
Query2: "get_information(relation='release_year', head_entity='output_of_query1')"

### USER:
Question: the movies written by [John Alan Schwartz] starred who
### ASSISTANT:
Step1: Find the movies that [John Alan Schwartz] written
Query1: "get_information(relation='written_by', tail_entity='John Alan Schwartz')"
Step2: Find actor of output_of_query1
Query2: "get_information(relation='starred_actors', head_entity='output_of_query1')"
Step3: Exclude [John Alan Schwartz] from output_of_query2
Query3: "set_difference(set1='output_of_query2', set2='John Alan Schwartz')"

### USER:
Question: who are the directors of the films written by [Laura Kerr]
### ASSISTANT:
Step1: Find the movies that [Laura Kerr] written
Query1: "get_information(relation='written_by', tail_entity='Laura Kerr')"
Step2: Find director of output_of_query1
Query2: "get_information(relation='directed_by', head_entity='output_of_query1')"
Step3: Exclude [Laura Kerr] from output_of_query2
Query3: "set_difference(set1='output_of_query2', set2='Laura Kerr')"

### USER:
Question: the screenwriter [Oscar Wilde] wrote films together with who
### ASSISTANT:
Step1: Find the movies that [Oscar Wilde] has written
Query1: "get_information(relation='written_by', tail_entity='Oscar Wilde')"
Step2: Find writer of output_of_query1
Query2: "get_information(relation='written_by', head_entity='output_of_query1')"
Step3: Exclude [Oscar Wilde] from output_of_query2
Query3: "set_difference(set1='output_of_query2', set2='Oscar Wilde')"

### USER:
Question: what types are the films written by [Stan Winston]
### ASSISTANT:
Step1: Find the movies that [Stan Winston] has written
Query1: "get_information(relation='written_by', tail_entity='Stan Winston')"
Step2: Find genre of output_of_query1
Query2: "get_information(relation='has_genre', head_entity='output_of_query1')"
\end{lstlisting}
\paragraph{LLM Prompt} LLM Prompt is the description for conditional graph query, following is the detail.

\begin{lstlisting}[basicstyle = \scriptsize\ttfamily, breaklines = true]
### SYSTEM:
Now you are learning to write conditional graph queries for natural language questions. 
The conditional graph query function is:
def get_information(relation, tail_entity, head_entity):
"""
    This function retrieves information by querying a data source using a given relation and tail entity as search criteria.
    Args:
        relation (str): The relation to the query that matches the tail_entity or contains the  head_entity which includes {'directed_by', 'has_imdb_rating', 'release_year', 'in_language', 'written_by', 'has_tags', 'has_imdb_votes', 'has_genre', 'starred_actors'}.
        tail_entity (str): The tail entity associated with the relation.
        head_entity (str): The head entity belongs to the relation.
    Returns:
        list: A list of query results.
    """
To finish the question correctly, you can the Set functions.
Set functions:
- set_difference(set1, set2)
Note: 
1. In get_information function, tail_entity and head_entity cannot occur together.
\end{lstlisting}
\underline{\textbf{MetaQA 3-hop}}: The prompt for MetaQA 3-hop has 11 demonstrations and the description for conditional graph query, following is the detail.
\paragraph{Initial Demonstrations}
MetaQA 3-hop has 11 demonstrations, the following is the detail.

\begin{lstlisting}[basicstyle = \scriptsize\ttfamily, breaklines = true]

### USER:
Question: who is listed as director of the films starred by [Titanic Town] actors
### ASSISTANT:
Step1: Find the actors of [Titanic Town]
Query1: "get_information(relation='starred_actors', head_entity='Titanic Town')"
Step2: Find movies that output_of_query1 acted in
Query2: "get_information(relation='starred_actors', tail_entity='output_of_query1')"
Step3: Exclude [Titanic Town] from output_of_query2
Query3: "set_difference(set1='output_of_query2', set2='Titanic Town')"
Step4: Find director of output_of_query3
Query4: "get_information(relation='directed_by', head_entity='output_of_query3')"

### USER:
Question: the films that share actors with the film [The Harvey Girls] were in which genres
### ASSISTANT:
Step1: Find the actors of [The Harvey Girls]
Query1: "get_information(relation='starred_actors', head_entity='The Harvey Girls')"
Step2: Find movies that output_of_query1 acted in
Query2: "get_information(relation='starred_actors', tail_entity='output_of_query1')"
Step3: Exclude [The Harvey Girls] from output_of_query2
Query3: "set_difference(set1='output_of_query2', set2='The Harvey Girls')"
Step4: Find genre of output_of_query3
Query4: "get_information(relation='has_genre', head_entity='output_of_query3')"

### USER:
Question: the movies that share actors with the movie [The Great Santini] were in which languages
### ASSISTANT:
Step1: Find the actors of [The Great Santini]
Query1: "get_information(relation='starred_actors', head_entity='The Great Santini')"
Step2: Find movies that output_of_query1 acted in
Query2: "get_information(relation='starred_actors', tail_entity='output_of_query1')"
Step3: Exclude [The Great Santini] from output_of_query2
Query3: "set_difference(set1='output_of_query2', set2='The Great Santini')"
Step4: Find language of output_of_query3
Query4: get_information(relation='in_language', head_entity='output_of_query3')"

### USER:
Question: when did the movies release whose actors also appear in the movie [Bigger Than Life]
### ASSISTANT:
Step1: Find the actors of [Bigger Than Life]
Query1: "get_information(relation='starred_actors', head_entity='Bigger Than Life')"
Step2: Find movies that output_of_query1 acted in
Query2: "get_information(relation='starred_actors', tail_entity='output_of_query1')"
Step3: Exclude [Bigger Than Life] from output_of_query2
Query3: "set_difference(set1='output_of_query2', set2='Bigger Than Life')"
Step4: Find release year of output_of_query3
Query4: "get_information(relation='release_year', head_entity='output_of_query3')"

### USER:
Question: who acted in the movies directed by the director of [Some Mother's Son]
### ASSISTANT:
Step1: Find the director of [Some Mother's Son]
Query1: "get_information(relation='directed_by', head_entity='Some Mother's Son')"
Step2: Find movies that output_of_query1 directed
Query2: "get_information(relation='directed_by', tail_entity='output_of_query1')"
Step3: Exclude [Some Mother's Son] from output_of_query2
Query3: "set_difference(set1='output_of_query2', set2='Some Mother's Son')"
Step4: Find actor of output_of_query3
Query4: "get_information(relation='starred_actors', head_entity='output_of_query3')"

### USER:
Question: what are the languages spoken in the films whose directors also directed [Police]
### ASSISTANT:
Step1: Find the director of [Police]
Query1: "get_information(relation='directed_by', head_entity='Police')"
Step2: Find movies that output_of_query1 directed
Query2: "get_information(relation='directed_by', tail_entity='output_of_query1')"
Step3: Exclude [Police] from output_of_query2
Query3: "set_difference(set1='output_of_query2', set2='Police')"
Step4: Find language of output_of_query3
Query4: "get_information(relation='in_language', head_entity='output_of_query3')"

### USER:
Question: the films that share directors with the films [Following] are written by who
### ASSISTANT:
Step1: Find the director of [Following]
Query1: "get_information(relation='directed_by', head_entity='Following')"
Step2: Find movies that output_of_query1 directed
Query2: "get_information(relation='directed_by', tail_entity='output_of_query1')"
Step3: Exclude [Following] from output_of_query2
Query3: "set_difference(set1='output_of_query2', set2='Following')"
Step4: Find writer of output_of_query3
Query4: "get_information(relation='written_by', head_entity='output_of_query3')"

### USER:
Question: who starred films for the screenwriter of [Deathtrap]
### ASSISTANT:
Step1: Find the writer of [Deathtrap]
Query1: "get_information(relation='written_by', head_entity='Deathtrap')"
Step2: Find movies that output_of_query1 write
Query2: "get_information(relation='written_by', tail_entity='output_of_query1')"
Step3: Exclude [Deathtrap] from output_of_query2
Query3: "set_difference(set1='output_of_query2', set2='Deathtrap')"
Step4: Find actor of output_of_query3
Query4: "get_information(relation='starred_actors', head_entity='output_of_query3')"

### USER:
Question: who directed the movies written by the writer of [Millennium Actress]
### ASSISTANT:
Step1: Find the writer of [Millennium Actress]
Query1: "get_information(relation='written_by', head_entity='Millennium Actress')"
Step2: Find movies that output_of_query1 write
Query2: "get_information(relation='written_by', tail_entity='output_of_query1')"
Step3: Exclude [Millennium Actress] from output_of_query2
Query3: "set_difference(set1='output_of_query2', set2='Millennium Actress')"
Step4: Find director of output_of_query3
Query4: "get_information(relation='directed_by', head_entity='output_of_query3')"

### USER:
Question: what types are the movies written by the writer of [Broadway Serenade]
### ASSISTANT:
Step1: Find the writer of [Broadway Serenade]
Query1: "get_information(relation='written_by', head_entity='Broadway Serenade')"
Step2: Find movies that output_of_query1 write
Query2: "get_information(relation='written_by', tail_entity='output_of_query1')"
Step3: Exclude [Broadway Serenade] from output_of_query2
Query3: "set_difference(set1='output_of_query2', set2='Broadway Serenade')"
Step4: Find genre of output_of_query3
Query4: "get_information(relation='has_genre', head_entity='output_of_query3')"

### USER:
Question: the movies that share writers with the movie [The Wedding Party] were released in which years
### ASSISTANT:
Step1: Find the writer of [The Wedding Party]
Query1: "get_information(relation='written_by', head_entity='The Wedding Party')"
Step2: Find movies that output_of_query1 write
Query2: "get_information(relation='written_by', tail_entity='output_of_query1')"
Step3: Exclude [The Wedding Party] from output_of_query2
Query3: "set_difference(set1='output_of_query2', set2='The Wedding Party')"
Step4: Find release year of output_of_query3
Query4: "get_information(relation='release_year', head_entity='output_of_query3')"

\end{lstlisting}
\paragraph{LLM Prompt} LLM Prompt is the description for conditional graph query, following is the detail.

\begin{lstlisting}[basicstyle = \scriptsize\ttfamily, breaklines = true]
### SYSTEM:
Now you are learning to write conditional graph queries for natural language questions. 
The conditional graph query function is:
def get_information(relation, tail_entity, head_entity):
"""
    This function retrieves information by querying a data source using a given relation and tail entity as search criteria.
    Args:
        relation (str): The relation to the query that matches the tail_entity or contains the  head_entity which includes {'directed_by', 'has_imdb_rating', 'release_year', 'in_language', 'written_by', 'has_tags', 'has_imdb_votes', 'has_genre', 'starred_actors'}.
        tail_entity (str): The tail entity associated with the relation.
        head_entity (str): The head entity belongs to the relation.
    Returns:
        list: A list of query results.
    """
To finish the question correctly, you can the Set functions.
Set functions:
- set_difference(set1, set2)
Note: 
1. In get_information function, tail_entity and head_entity cannot occur together.
\end{lstlisting}

\subsection{CronQuestion}\label{sec:cron_prompt}
\paragraph{Initial Demonstrations}
CronQuestion has 8 demonstrations, the following is the detail.
\begin{lstlisting}[basicstyle = \scriptsize\ttfamily, breaklines = true]
### USER:
Question: EATCS award was awarded to Moshe Vardi during the year
Relations: ['award received']
Annotation: {'head': 'Moshe Vardi', 'tail': 'EATCS award'}
### ASSISTANT:
Step1: Get the time of the EATCS award received by Moshe Vardi
Query1: "get_information(head_entity='Moshe Vardi', relation='award received', tail_entity='EATCS award', key='time')"

### USER:
Question: In 1981 which award did Allan Edwall win
Relations: ['award received']
Annotation: {'head': 'Allan Edwall', 'time': '1981'}
### ASSISTANT:
Step1: Get the award received by Allan Edwall in 1981
Query1: "get_information(head_entity='Allan Edwall', relation='award received', key='time', value='1981')"

### USER:
Question: The first time Gordon Campbell, Baron Campbell of Croy had been Member of the 44th Parliament of the United Kingdom
Relations: ['position held']
Annotation: {'head': 'Gordon Campbell, Baron Campbell of Croy', 'tail': 'Member of the 44th Parliament of the United Kingdom', 'adj': 'first'}
### ASSISTANT:
Step1: Get the time/year of the Gordon Campbell, Baron Campbell of Croy position held Member of the 44th Parliament of the United Kingdom
Query1: "get_information(head_entity='Gordon Campbell, Baron Campbell of Croy', relation='position held', tail_entity='Member of the 44th Parliament of the United Kingdom', key='time')"
Step2: Get the min/first/earliest time/year of output_of_query1
Query2: "min(set='output_of_query1')"

### USER:
Question: Who's the last team Roberto Colombo played on
Relations: ['member of sports team']
Annotation: {'head': 'Roberto Colombo', 'adj': 'last'}
### ASSISTANT:
Step1: Get the time of Roberto Colombo is member of sports team
Query1: "get_information(head_entity='Roberto Colombo', relation='member of sports team', key='time')"
Step2: Get the last/max time output_of_query1
Query2: "max(set='output_of_query1')"
Step3: Get the name of team Roberto Colombo is member of sports team when the time is output_of_query2
Query3: "get_information(head_entity='Roberto Colombo', relation='member of sports team', key='time', value='output_of_query2')"

### USER:
Question: Who were the players playing in Sunderland A.F.C. with Chris Lumsdon?
Relations: ['member of sports team']
Annotation: {'head': 'Chris Lumsdon', 'tail': 'Sunderland A.F.C.'}
### ASSISTANT:
Step1: Get the time of Chris Lumsdon is member of sports team Sunderland A.F.C.
Query1: "get_information(head_entity='Chris Lumsdon', relation='member of sports team', tail_entity='Sunderland A.F.C.', key='time')"
Step2: Get the players is member of sports team Sunderland A.F.C. when the time is output_of_query1
Query2: "get_information(relation='member of sports team', tail_entity='Sunderland A.F.C.', key='time', value='output_of_query1')"

### USER:
Question: Who held the position of state treasurer when Michael Matheson was the Member of the 3rd Scottish Parliament
Relations: ['position held']
Annotation: {'head': 'Michael Matheson', 'tail': 'Member of the 3rd Scottish Parliament'}, ['state treasurer']
### ASSISTANT:
Step1: Get the time of Michael Matheson position held Member of the 3rd Scottish Parliament
Query1: "get_information(head_entity='Michael Matheson', relation='position held', tail_entity='Member of the 3rd Scottish Parliament', key='time')"
Step2: Get the people position held is state treasurer when the time is output_of_query1
Query2: "get_information(relation='position held', tail_entity='state treasurer', key='time', value='output_of_query1')"

### USER:
Question: Who were Member of the Swiss Federal Council after 18th century
Relations: ['position held']
Annotation: {'tail': 'Member of the Swiss Federal Council', 'event_head': '18th century', 'type': 'after'}
### ASSISTANT:
Step1: To get the after, find the time of 18th century
Query1: "get_information(head_entity='18th century', key='time')"
Step2: Find the max of output_of_query1
Query2: "max(set='output_of_query1')"
Step3: Get the time of position held is Member of the Swiss Federal Council
Query3: "get_information(relation='position held', tail_entity='Member of the Swiss Federal Council', key='time')"
Step4: Get the time of output_of_query3 after output_of_query2
Query4: "keep(set='output_of_query3', value>'output_of_query2')"
Step5: Get the min time of output_of_query4
Query5: "min(set='output_of_query4')"
Step6: Get the name whose position held is Member of the Swiss Federal Council when the time is output_of_query5
Query6: "get_information(relation='position held', tail_entity='Member of the Swiss Federal Council', key='time', value='output_of_query5')"

### USER:
Question: Which team did Michael Gray play for before England national football team
Relations: ['member of sports team']
Annotation: {'head': 'Michael Gray', 'tail': 'England national football team', 'type': 'before'}
### ASSISTANT:
Step1: To get the before, find the time of Michael Gray is member of sports team England national football team
Query1: "get_information(head_entity='Michael Gray', relation='member of sports team', tail_entity='England national football team', key='time')"
Step2: Find the min of output_of_query1
Query2: "min(set='output_of_query1')"
Step3: Get the time of Michael Gray is member of sports team
Query3: "get_information(head_entity='Michael Gray', relation='member of sports team', key='time')"
Step4: Get the time of output_of_query3 before output_of_query2
Query4: "keep(set='output_of_query3', value<output_of_query2)"
Step5: Get the max time of output_of_query4
Query5: "max(set='output_of_query4')"
Step6: Get the name of team Michael Gray is member of sports team when the time is output_of_query5
Query6: "get_information(head_entity='Michael Gray', relation='member of sports team', key='time', value='output_of_query5')"

\end{lstlisting}
\paragraph{LLM Prompt} LLM Prompt is the description for conditional graph query, following is the detail.

\begin{lstlisting}[basicstyle = \scriptsize\ttfamily, breaklines = true]
### SYSTEM:
Now you are learning to write conditional graph queries for natural language questions. 
And conditional graph query function is:
def get_information(relation, tail_entity, head_entity, key, value):
"""
    This function retrieves information by querying a data source using a given relation and tail entity as search criteria.
    Args:
        relation (str): The relation to the query that matches the tail_entity or head_entity.
        tail_entity (str): The tail entity associated with the relation.
        head_entity (str): The head entity belongs to the relation.
        key (str):  The key to query that matches with time.
        value (str):  The value associated with or belongs to the key.
    Returns:
        list: A list of query results.
    """
Note: you need to use keep(set, value) to filter the results for before or after question.
Your task is to break down the original natural language question step by step with given relations and annotations. Following are some examples.
\end{lstlisting}

\section{LLM function prompt}\label{sec:LLM_prompt}
In LLM function we optimize the following three kinds of problems, including: multiple choice problems, extreme value problems and comparison problems.

A multiple-choice question means that the answer is chosen according to the description of the question. The prompt is designed as follows:

\begin{lstlisting}[basicstyle = \scriptsize\ttfamily, breaklines = true]
### SYSTEM:
Answer questions based on the information provided.
### USER:
Question: who won more silvers, venezuela or chile?
Information provided:
Step1: Find the Nation of Venezuela. output_of_query1: {'[line_1]'}
Step2: Find the Nation of Chile. output_of_query2: {'[line_4]'}
Step3: Find the number of silvers of output_of_query1. output_of_query3: {'8'}
Step4: Find the number of silvers of output_of_query2. output_of_query4: {'4'}
Step5: Compare the number of silvers of output_of_query3 and output_of_query4.
Just tell me the answer in a word or phrase without spaces or newlines."
### ASSISTANT:
Venezuela
\end{lstlisting}

Extremum and comparison problems are those where you want to compare more than just numbers. You might want to compare units (1 km), special representations (1-9), etc. 

The \textbf{Extremum problem} prompt is designed as follows:

\begin{lstlisting}[basicstyle = \scriptsize\ttfamily, breaklines = true]
### SYSTEM:
You are a helpful assistant. If the data format provided is complex, use common sense.
Please enclose the final answer in quotation marks for ease of extraction.
### USER:
Question: Get the min output_of_query1 And Output_query1: {'12.15 (87)', '14.16 (100)', '8.5 (53)', '10.9 (69)', '6.7 (43)', '2.9 (21)', '11.12 (78)', '8.9 (57)', '9.11 (65)', '13.6 (84)', '11.8 (74)', '9.4 (58)', '14.11 (95)', '14.14 (98)', '11.2 (68)', '4.11 (35)', '19.14 (128)', '17.13 (115)', '8.10 (58)', '17.6 (108)', '6.9 (45)', '11.11 (77)'}.
Please tell me the only data who satisfies the condition.
### ASSISTANT:
Based on the given data, it appears that the min data is '2.9 (21)'.
### USER:
Get the best position of output_of_query1 And Output_query1: {'13th (q)', '5th', '7th', '-'}.
Please tell me the only data who satisfies the condition.
### ASSISTANT:
Based on the given data, it appears that the best data is '5th'.
\end{lstlisting}

The \textbf{Comparison problem} prompt is designed as follows:
\begin{lstlisting}[basicstyle = \scriptsize\ttfamily, breaklines = true]
### SYSTEM:
You are a helpful assistant.
### USER:
Question: Find the players who are taller than output_of_query2 And Output_query1: {'[line_8]'}.Output_query2: {'6-4'}.
All data are given in the format of (data, line_number):
{('6-3', '[line_3]'), ('6-9', '[line_6]'), ('6-8', '[line_1]'), ('6-5', '[line_4]'), ('6-5', '[line_7]'), ('5-11', '[line_5]'), ('5-10', '[line_2]')}.
Please check it step by step and tell me the line_number whose data satisfies the condition.
### ASSISTANT:
To find the players who are taller than '6-4', we need to check the values given in the format ('height', 'line_number') one by one.
Here are the step-by-step results:
1. ('6-3', '[line_3]') - Not taller than '6-4'.
2. ('6-9', '[line_6]') - Taller than '6-4'.
3. ('6-8', '[line_1]') - Taller than '6-4'.
4. ('6-5', '[line_4]') - Taller than '6-4'.
5. ('6-5', '[line_7]') - Taller than '6-4'.
6. ('5-11', '[line_5]') - Not taller than '6-4'.
7. ('5-10', '[line_2]') - Not taller than '6-4'.
The line numbers whose heights satisfy the condition (taller than '6-4') are [line_6], [line_1], [line_4], and [line_7].
### USER:
Find the line_number which has more strokes than output_of_query2 And Output_query1: {'[line_3]'}.Output_query2: {'3 strokes'}.
All data are given in the format of (data, line_number):
{('1 stroke', '[line_5]'), ('Playoff', '[line_2]'), ('5 strokes', '[line_4]'), ('1 stroke', '[line_1]'), ('2 strokes', '[line_2]')}
Please check it step by step and tell me the line_number whose data satisfies the condition.
### ASSISTANT:
To find the line_number which has more strokes than {'3 strokes'}, we need to check the values given in the format ('strokes', 'line_number')  one by one.
1. ('1 stroke', '[line_5]') - Less strokes than '3 strokes'.
2. ('Playoff', '[line_2]') - Not applicable for stroke comparison.
3. ('5 strokes', '[line_4]') - More strokes than '3 strokes'.
4. ('1 stroke', '[line_1]') - Less strokes than '3 strokes'.
5. ('2 strokes', '[line_2]') - Less strokes than '3 strokes'.
The line number(s) whose number of strokes satisfies the condition (more than '3 strokes') is: [line_4]."

### USER:
Find the positions that are at least 10th place or better And 
All data are given in the format of (data, line_number):
{('13th (q)', '[line_1]'), ('5th', '[line_3]'), ('7th', '[line_2]'), ('-', '[line_4]')}.
Please check it step by step and tell me the line_number whose data satisfies the condition."
### ASSISTANT:
To find the positions that are at least 10th place or better, we need to compare the positions given in the format ('position', 'line_number') with the condition of being 10th place or better. Here are the step-by-step results:
1. ('13th (q)', '[line_1]') - Not 10th place or better.
2. ('5th', '[line_3]') - 10th place or better.
3. ('7th', '[line_2]') - 10th place or better.
4. ('-', '[line_4]') - Not applicable for position comparison.
The line numbers whose positions satisfy the condition (10th place or better) are: [line_2], and [line_3]."
\end{lstlisting}

\end{document}